\PassOptionsToPackage{table,xcdraw}{xcolor}
\documentclass{rdm}

\usepackage{lineno}
\usepackage{hyperref}
\usepackage{cleveref}
\crefname{figure}{Fig.}{Figs.}
\crefname{table}{Tab.}{Tabs.}
\crefname{equation}{Eq.}{Eqs.}
\crefname{section}{Sec.}{Secs.}

\usepackage{amsmath}
\usepackage{amssymb}
\usepackage{mathtools}
\usepackage{algorithm}
\usepackage{multicol}
\usepackage{multirow}
\usepackage{wrapfig}
\usepackage{needspace}
\usepackage{indentfirst}
\usepackage{algorithmicx}
\usepackage{algpseudocode}%
\usepackage[table,xcdraw]{xcolor}
\usepackage{xcolor}
\usepackage{soul}

\newcommand{\IndState}[1][1]{\State\hspace{#1cm}}
\newcommand{\IndStatex}[1][1]{\State\hspace{#1cm}}

\newcommand{\bstcell}{\cellcolor[HTML]{ffc7a2}}
\newcommand{\midcell}{\cellcolor[HTML]{f2e8c3}}
\newcommand{\worcell}{\cellcolor[HTML]{d0d0d0}}
\definecolor{lightred}{HTML}{ffc7a2}
\definecolor{lightorange}{HTML}{f2e8c3}
\definecolor{lightyellow}{HTML}{d0d0d0}

\newcommand{\bsttxt}[1]{\sethlcolor{lightred}\hl{#1}}
\newcommand{\midtxt}[1]{\sethlcolor{lightorange}\hl{#1}}
\newcommand{\wortxt}[1]{\sethlcolor{lightyellow}\hl{#1}}

\newcommand{\shortmethod}{RDM\xspace}

\title{RDM: Recurrent Diffusion Model for Human Motion Generation}

\addauthor{Mirgahney Mohamed}{m.h.mohamed.20@ucl.ac.uk}{1}
\addauthor{Harry Jake Cunningham}{jake.cunningham.21@ucl.ac.uk}{1}
\addauthor{Riccardo Barbano}{riccardo.barbano.19@ucl.ac.uk}{1}
\addauthor{Marc Peter Deisenroth}{m.deisenroth@ucl.ac.uk}{1}
\addauthor{Lourdes Agapito}{l.agapito@ucl.ac.uk}{1}

\addinstitution{
 Department of Computer Science\\
 University College London\\
 London, UK
}

\runninghead{Mohamed \rdmaEtAl}{Recurrent Diffusion Model}

\def\etal{\emph{et al}\rdmaOneDot}

\begin{document}

\maketitle

\begin{abstract}
Human motion generation is a challenging task due to its high dimensionality and the difficulty of generating fine-grained motions.  
Diffusion methods have been proposed due to their high sample quality and expressiveness. 
Early approaches treat the entire sequence as a whole, which is computationally expensive and restricts sequence length. 
In contrast, autoregressive diffusion models generate longer sequences. 
However, their reliance on fully denoising previous frames complicates training and inference. 
Consequently, we propose RDM, a new recurrent diffusion formulation inspired by Recurrent Neural Networks (RNNs).
RDM explicitly conditions the diffusion process on preceding noisy frames, avoiding the cost of full denoising. To overcome the challenge of maintaining its probabilistic nature, we employ Normalizing Flows to model recurrent connections.
Our evaluations demonstrate RDM's effectiveness: it achieves comparable performance to autoregressive baselines and generates long sequences that remain aligned with the text.
RDM also skips diffusion steps during inference, reducing computational cost by up to $18\times$ over autoregressive baselines.
Project page: \url{https://www.mirgahney.com/rfdm.io/}.
\end{abstract}
\section{Introduction}
\label{sec:intro}
Human motion generation is crucial in computer animation, with applications spanning gaming to robotics.
Despite notable progress, it remains challenging due to the vast space of possible motions and the sophisticated equipment required to acquire the data.
Therefore, automating motion generation from natural signals, such as text, offers an intuitive interface for users and significantly reduces the effort involved.

Variational Auto-Encoders (VAEs)~\cite{kingma2013auto} demonstrated success in mapping various modalities to motion~\cite{li2021learn, tseng2022edge, TEMOS_Petrovich2022, tevet2022motionclip}.
Nevertheless, VAEs inherently impose restrictions on the target distribution, limiting their expressiveness.

In contrast, diffusion models~\cite{pmlrv37sohldickstein15} do not impose restrictions on target distributions~\cite{tevet2023human} and exhibit excellent capacity for many-to-many distribution matching~\cite{NEURIPS2020_4c5bcfec}.
MotionDiffuse \cite{zhang2022motiondiffuse} and MDM \cite{tevet2023human} are pioneering works that employ diffusion methods \cite{pmlrv37sohldickstein15} for human motion generation. 
MLD \cite{chen2023executing} further advanced this by applying diffusion in latent space, demonstrating efficient training and inference.

\begin{figure}[t]
    \centering
    \includegraphics[width=\linewidth]{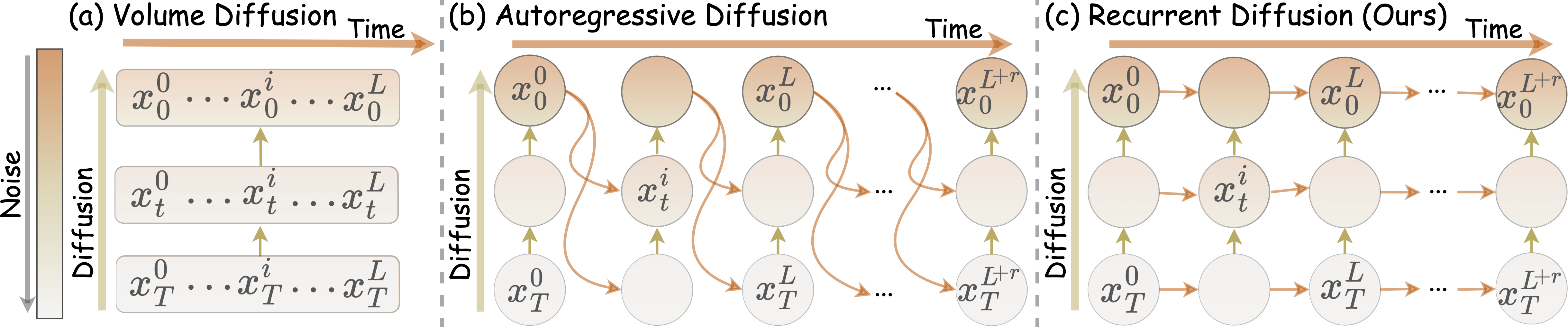}
    \caption{\textbf{Temporal diffusion models inference.}
    a) Volume diffusion treats the entire sequence as a monolithic input and diffuses it as a whole.
    b) Autoregressive diffusion conditions only the reverse process on previously estimated frames while ignoring the forward process.
    c) Recurrent diffusion leverages a recurrent formulation and explicitly conditions both forward and reverse processes on previous noisy frames.
    }
    \label{fig:teaser}
\end{figure}

However, these approaches \cite{zhang2022motiondiffuse, chen2023executing, tevet2023human}, which we term \textit{volume-diffusion}, treat the entire sequence as a monolithic input (\cref{fig:teaser} (a)).
Consequently, this monolithic treatment contributes to motion incoherence and makes handling variable or extended sequence lengths computationally prohibitive, owing to the high cost of diffusing an entire sequence at once.

\begin{figure}[t]
    \centering
    \includegraphics[width=\linewidth]{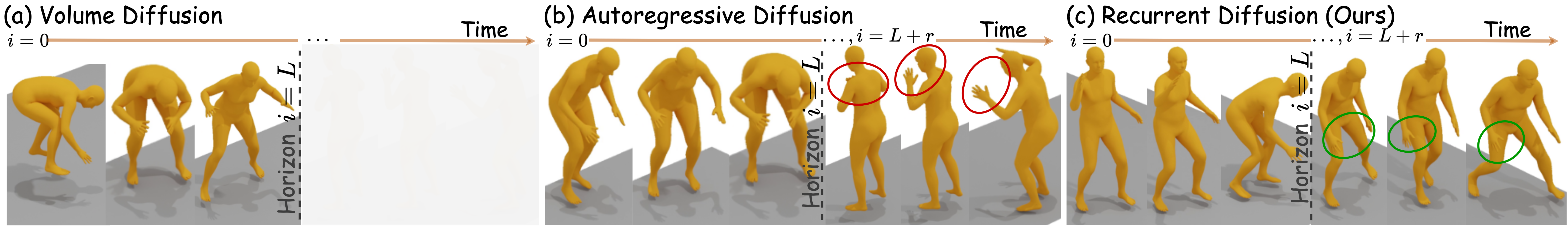}
    \caption{\textbf{Temporal diffusion generation.}
    Volume diffusion (a) generates sequences limited by the horizon ($L$), shown by the dashed line;
    On the other hand, autoregressive (b) and recurrent diffusion (c) extend sequences beyond the horizon ($L + r$).
    However, recurrent diffusion produces better-aligned sequences (green circles) with ``dribbling with a basketball''.}
    \label{fig:teaser_1}
\end{figure}

Two primary approaches emerged to address the challenge of generating motion via sequential segments.
TEDi \cite{zhang2024tedi} extends diffusion by incorporating a temporally varying noise level within the forward process while ignoring the reverse. 
Other methods \cite{ZhaoDart2025, shi2024amdm, meng2024rethinking, tevet2025closd, han2023amd} condition only the reverse process on previously estimated clean frames, ignoring the forward, which we term \textit{autoregressive-diffusion} (\cref{fig:teaser} (b)).
Although these approaches allow segmented generation, their autoregressive nature complicates training and inference, as it necessitates fully denoising preceding frames to generate future frames.

To overcome these limitations, we propose \textit{Recurrent Diffusion Model (RDM)}, a novel diffusion framework for generative temporal modelling.
RDM leverages a recurrent formulation, analogous to Recurrent Neural Networks (RNNs)~\cite{Rumelhart_86, osti_6910294}, extending sequences beyond the training horizon with better-aligned motion (\cref{fig:teaser,fig:teaser_1} (c)).
Within this recurrent formulation, both the forward (noise addition) and reverse (denoising) diffusion processes are explicitly conditioned on previous noisy frames.
This design entangles denoising and future frame prediction tasks, providing flexibility over segment lengths and counts.
Note that removing recurrence turns \shortmethod into a volume diffusion model. 
We will demonstrate that this ``recurrence'' improves results (\cref{tab:hml_metrics}) and significantly reduces computational complexity compared to baselines (\cref{tab:compute_cost}).

Although this recurrent formulation offers compelling properties, a key challenge is preserving its probabilistic nature. 
This is because the recurrent transformation does not inherently guarantee valid probability distributions.
Such deviations theoretically invalidate the diffusion model's loss, thereby compromising training and leading to incorrect samples.
To address this, we utilize normalizing flows~\cite{pmlr_rezende15, dinh2017density} to model temporal dependencies, thereby preserving the probability under recurrent transformations. 
Our contributions are:

\begin{itemize}
    \item A novel recurrent diffusion formulation that leverages Normalizing Flows to model spatiotemporal dependencies via noisy hidden states, establishing a non-Markovian framework for motion synthesis.
    
    \item A flexible generation mechanism that models sequential segments interactively, enabling adaptable temporal lengths without relying on monolithic volume diffusion.
    \item An efficient staircase strategy that reduces inference latency by skipping redundant diffusion steps, achieving up to $18\times$ speedup over autoregressive baselines while maintaining motion fidelity.
    
\end{itemize}
These merits are demonstrated in text-to-motion generation tasks on the KIT-ML~\cite{Plappert2016} and HumanML3D~\cite{Guo_2022_CVPR} datasets.

\section{Related Work}
\label{sec:rw}

\subsection{Conditioned Motion Generation}
Variational Auto-Encoders (VAEs)~\cite{kingma2013auto} have been widely used in human motion modelling.
Notable works include Motion Transformation VAE (MT-VAE)~\cite{Yan2018}, which decomposes motion into atomic units with an LSTM encoder-decoder, and Aliakbarian \etal~\cite{Aliakbarian_2020_CVPR}, which integrates root variations with previous poses to reduce noise. 
Recently, MotionCLIP~\cite{tevet2022motionclip} introduces a transformer-based motion autoencoder, whereas AvatarCLIP~\cite{Hong2022} employs a volume rendering model for geometry and texture generation. 
TEMOS~\cite{TEMOS_Petrovich2022}, similar to ACTOR~\cite{petrovich21actor}, learns joint motion and text spaces with transformers.
Lastly, MoMask~\cite{guo2024momask} models human motion as hierarchical tokens, with a Masked Transformer predicting base tokens from text and filling sequences during generation. 

Despite their plausible results, VAE-based methods are often limited in their expressiveness due to inherent constraints imposed on the learned distributions.

In contrast to VAEs, Normalizing Flows (NFs)~\cite{pmlr_rezende15} train efficiently with exact maximum likelihood. 
Inspired by GLOW~\cite{Kingma_NEURIPS2018}, MoGlow~\cite{henter2020moglow} proposes an autoregressive normalization network for motion sequences. 
Normalizing flows, however, are limited in their capacity to model complex distributions~\cite{NEURIPS2020_ecb9fe2f, NEURIPS2020_41d80bfc}.

Generative Adversarial Networks (GANs)~\cite{Goodfellow2014_gan} have been explored for motion generation. 
HP-GAN~\cite{BarsoumCVPRW2018} uses a sequence-to-sequence model for probabilistic human motion prediction with a custom loss. 
Harvey \etal~\cite{Harvey2020} address motion blurriness based on adversarial recurrent networks. 
Additionally, Wang \etal~\cite{Wang2020} decouples the generator into a smooth latent transition and a global skeleton decoder for individual latent frames. 
Although GANs do not impose explicit constraints on the target distribution, they are hard to train and often suffer from mode collapse~\cite{Durall2020CombatingMC, Youssef_2022}.

\subsection{Diffusion Methods for Motion Generation}
Inspired by non-equilibrium thermodynamics, Sohl-Dickstein \etal~\cite{pmlrv37sohldickstein15} propose a Markov chain that gradually adds noise to data (\textit{forward}) and learns to recover it with a neural network (\textit{reverse}).
With the success of diffusion models~\cite{NEURIPS2020_4c5bcfec, rombach2022high}, research shifted to temporal data such as videos~\cite{ho2022video, chen2023controlavideo}, trajectory planning~\cite{NEURIPS2024_2aee1c41}, and human motion~\cite{zhang2022motiondiffuse, chen2023executing}.

\noindent \textbf{Volume diffusion.}
MotionDiffuse~\cite{zhang2022motiondiffuse} is among the first to apply diffusion to text-to-motion generation using transformer generator networks on the SMPL pose space~\cite{SMPL_2015}. 
MLD~\cite{chen2023executing} and MotionMamba~\cite{zhang2025motion} learn a latent space via a transformer VAE and apply diffusion within it, using a transformer denoiser and a Mamba-based~\cite{mamba2} SSM denoiser, respectively.
Using diffusion in the latent space enables them to train and sample orders of magnitude faster. 
However, this heavily depends on the quality of the latent space learned by the VAE, which requires careful tuning and regularization and can produce artefacts~\cite{hoogeboom2024simpler}. 
Recently, Light-T2M~\cite{light_t2m_2025} exploits local and global sequence information via an efficient Mamba denoiser, achieving SOTA performance at only $10\%$ of MoMask's~\cite{guo2024momask} size.

Despite their success, these approaches generate temporal predictions limited to a short horizon~\cite{zhang2022motiondiffuse, chen2023executing}.

\noindent \textbf{Autoregressive diffusion.}
TEDi~\cite{zhang2024tedi} extends DDPM to the temporal dimension through temporally varying denoising, generating motion frames during inference and feeding in newly noised frames.
AMD~\cite{han2023amd} employs an autoregressive inverse process, an early effort to integrate temporal constraints into diffusion.
DART~\cite{ZhaoDart2025} learns motion primitives in the latent space while UniPhys~\cite{wu2025uniphys} encodes variable motion frames into the latent space.
Meng \etal~\cite{meng2024rethinking} retain 3D features to ensure a uniform distribution before autoregressively denoising with a masked transformer.
Meanwhile, CloSD~\cite{tevet2025closd} and A-MDM~\cite{shi2024amdm} apply the same autoregressive conditioning in input pose space rather than latent space.

As discussed in \cref{sec:intro}, these autoregressive diffusion methods~\cite{ZhaoDart2025, shi2024amdm, meng2024rethinking, tevet2025closd, han2023amd} condition only the reverse process on previously estimated frames, inheriting the same training and inference complications.

Inspired by recurrent networks~\cite{Rumelhart_86, osti_6910294}, we developed our \textit{Recurrent Diffusion Model (RDM)} for temporal modelling, ensuring both forward and reverse processes follow temporal constraints by autoregressively adding and removing noise, departing from previous diffusion models~\cite{zhang2022motiondiffuse, chen2023executing, ZhaoDart2025, shi2024amdm, meng2024rethinking, tevet2025closd, han2023amd}.
Unlike Diffusion-Forcing~\cite{NEURIPS2024_2aee1c41}, the added noise is correlated, enabling the generation of sequences by sampling only the first segment.
By leveraging normalizing flows~\cite{pmlr_rezende15, dinh2017density, Kingma_NEURIPS2018} to model temporal transitions, our method captures more complex distributions than simple Gaussians, boosting capacity. 
RDM also skips diffusion steps during inference, reducing computational costs.

\begin{figure*}[t]
    \centering
    \includegraphics[width=\linewidth]{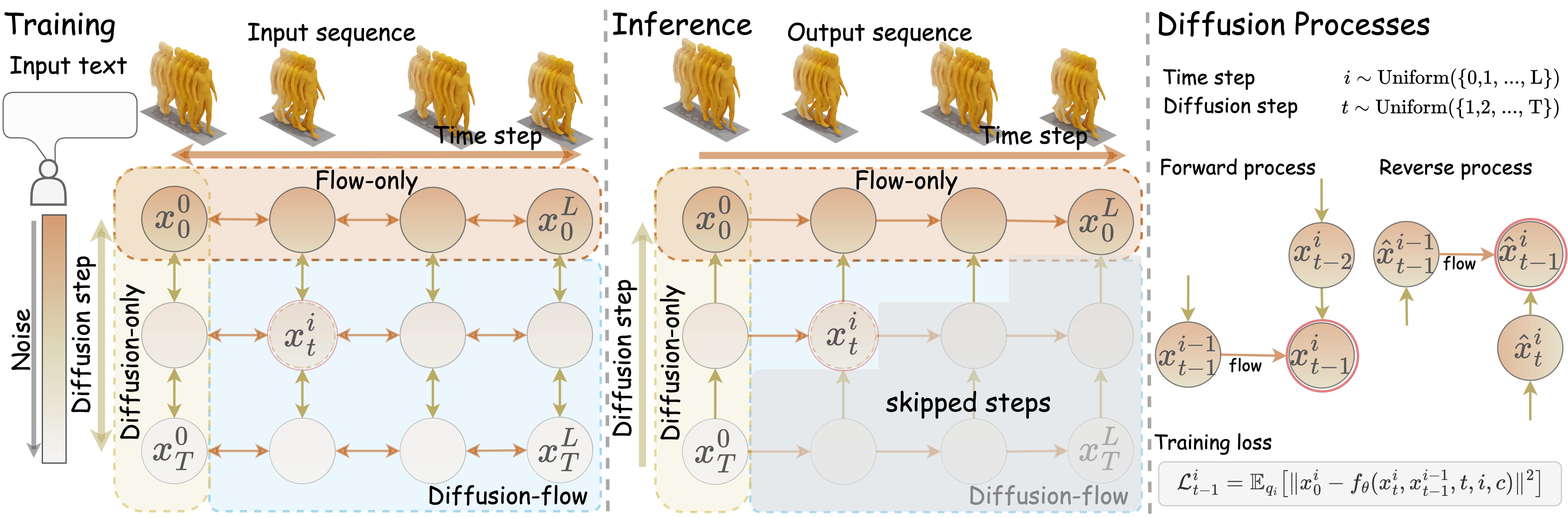}
    \caption{
    \textbf{Method overview:}
    RDM extends diffusion to temporal sequences via a 2D dependency grid across diffusion steps $t$ and temporal steps $i$, using invertible Normalizing Flows to keep the recurrent transformations probabilistically tractable.
    Three regimes compose the grid: \textit{flow-only} ($t=0$) transforms between clean segments; \textit{diffusion-only} ($i=0$) is standard diffusion on the first segment; \textit{diffusion-flow} ($t{>}0, i{>}0$) jointly conditions on the preceding noisy state $x^i_{t-1}$ and the previous segment's diffused state $x^{i-1}_t$ (see \cref{sec:method} for the full formulation).
    }
    \label{fig:rfdm_arch}
\end{figure*}

\section{Background}
\label{sec:back}

\subsection{Diffusion Models}
\label{back:diff}
Diffusion models~\cite{pmlrv37sohldickstein15, NEURIPS2020_4c5bcfec} add noise to data via a \textit{forward process} and recover it through a \textit{reverse process}, both following a Markov chain.
Given a pair $(x^{(j)}, u^{(j)})$, where $x^{(j)}$ is a sequence of poses and $u^{(j)}$ the description, each $x^{(j)}$ has $F$ frames $(H^{i \; (j)}) \; i \in \{1,2,...,F\}$, with $H^{i \; (j)} \in \mathbb{R}^D$ as the pose of the $i$-th frame. 
The reverse process $p_{\theta}(x_{t-1}| x_t)$ is modelled as a Gaussian transition learned with a neural network;
\begin{equation}
    \begin{aligned}
        p_{\theta}(x_{0:T}) &:= p(x_T) \prod_{t=1}^T p_{\theta}(x_{t-1}| x_t) \\
        p_{\theta}(x_{t-1}| x_t) &:= \mathcal{N}(x_{t-1}; \mu_{\theta}(x_t, t, c), \Sigma)
    \end{aligned}
\end{equation}
Here, $\mu_{\theta}$ is parameterized by neural networks, while $\Sigma$ can be either learned jointly or held fixed.
Here, $c$ denotes textual description embeddings extracted with a pretrained neural network.
The approximate posterior distribution $q(x_t \mid x_{t-1})$ of the diffusion model is called the \textit{forward process} or \textit{diffusion process}. 
This process gradually adds Gaussian noise to the data according to a variance scheduler $\beta_1, \beta_2, ..., \beta_T$, which can be learned \cite{pmlrv37sohldickstein15} or fixed:
\begin{equation}
    \begin{aligned}
        q(x_{1:T}| x_0) &= \prod_{t=1}^T q(x_t| x_{t-1}) \\
        q(x_t| x_{t-1}) &:= \mathcal{N}(x_t; \sqrt{1 - \beta_t} x_{t-1}, \beta_t \mathbf{I})
    \end{aligned}
\end{equation}
Training is performed by optimizing the Kullback-Leibler (KL) divergence between the forward and reverse processes.
As suggested by Ho \etal~\cite{NEURIPS2020_4c5bcfec}, choosing a specific parametrization of $\mu_\theta(x_t,t,c)$, yields the loss:
\begin{equation}\label{eq:diff_loss}
    \begin{aligned}
        \mathcal{L}_{\theta} &= \mathbb{E}_{q} \left[ w(t) \|x - f_{\theta}(x_t,t, c)\|^2\right]
    \end{aligned}
\end{equation}
Here, $f_{\theta}(x_t,t, c)$ is a learnable neural network, and the weight $w(t)$ can be freely specified.
Efficient training can be achieved using Monte Carlo integration to approximate the expectation. 
We uniformly sample the diffusion step $t$ and minimize the loss using stochastic gradient descent.

\subsection{Normalizing Flows}
\label{back:flow}
A normalizing flow~\cite{pmlr_rezende15} is a transformation $Y = f(X)$ that transforms one random variable $X$ into another $Y$ via successive volume-preserving, invertible and bijective $K$ transformations $f_K \circ f_{K-1} \circ ... \circ f_1$ (\cref{eq:flow}).

We use normalizing flows for modelling temporal human motion. 
Given a data point $x^{(j)}$ with $F$ frames from a distribution $q(\mathcal{X})$, we split it into $L$ segments $\{x^i\}_{i=0}^L$, each with equal length. 
Normalizing flow predicts the next segment distribution $x^{i+1} \sim P_{X^{i+1}}$ from the previous $x^i \sim P_{X^{i}}$. 
Applying the change of variable theorem, we obtain:
\begin{align}
    x^{i+1} &= x_K = f_K \circ f_{K-1} \circ ... \circ f_1(x^i) 
    \label{eq:flow}
    \\
    P_{X^{i+1}}(x^{i+1}) &= P_{X^{i}}(x^{i})\prod_{k=1}^{K}\left|\det \left(\frac{\partial f_k(x_{k-1})}{\partial x_{k-1}}\right) \right|^{-1}
\end{align}
Samples from the transformed distribution $P_{X^{i+1}}$ are generated by sampling from the base distribution $x^{i} \sim P_{X^{i}}$, then applying the transformations in \cref{eq:flow} to get $x^{i+1}$.
\begin{figure*}[t] 
    \centering
    \includegraphics[width=\linewidth]{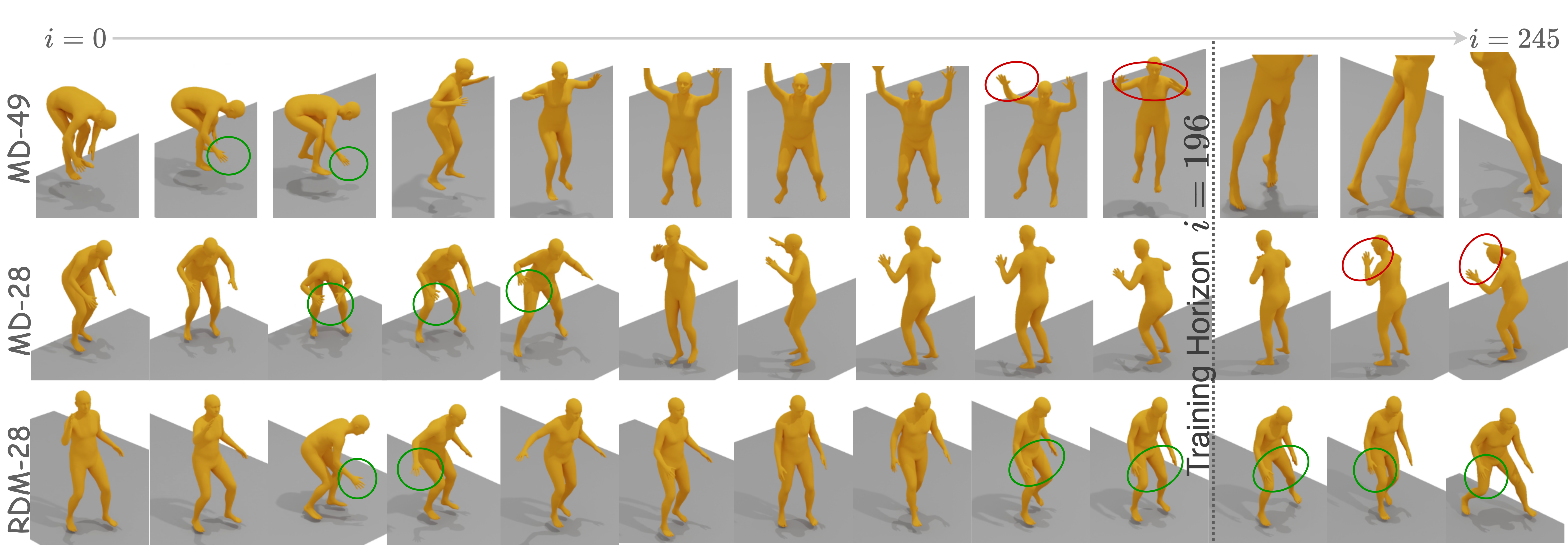}
    \caption{
    \textbf{Qualitative results on extended generation on the HumanML3D dataset~\cite{Guo_2022_CVPR}.}
    We evaluate extended generation with the prompt ``A person is dribbling with a basketball'', generating motion up to $245$ frames (right of the dashed line); green/red circles mark correct/broken dribbling frames.
    Within the training horizon (left of the dashed line), MD-49 dribbles correctly at first but loses the motion near the boundary, raising its arms overhead, and fails beyond the horizon.
    MD-28 dribbles correctly for longer, but its hands stop dribbling once past the horizon.
    In contrast, \shortmethod-28 keeps dribbling correctly throughout.
    }
    \label{fig:skeleton_figs}
\end{figure*}

\section{Recurrent Diffusion Model}
\label{sec:method}

We extend diffusion models~\cite{pmlrv37sohldickstein15, NEURIPS2020_4c5bcfec} into the temporal dimension by establishing a recurrent diffusion framework. 
Specifically, we structure the generative process over a two-dimensional dependency grid across diffusion and temporal steps (\cref{fig:rfdm_arch}).

Formally, let $i$ denote the temporal step and $t$ the diffusion step.
As shown in~\cref{fig:rfdm_arch}, the diffusion state $x^i_t$ is jointly conditioned on the preceding diffusion state within the same temporal segment, $x^i_{t-1}$ (intra-segment denoising), and the corresponding diffusion state from the previous temporal segment, $x^{i-1}_t$ (inter-segment recurrence). 
This formulation enforces global temporal continuity across sequential segments while preserving fine-grained motion within each segment.

However, a critical challenge arises: recurrent conditioning on intermediate noisy frames does not inherently guarantee valid probability distributions. 
This mathematically invalidates the standard diffusion training objective (\cref{eq:diff_loss}), as the KL divergence between the forward and reverse processes becomes intractable or undefined, leading to degraded sample quality during inference.
To resolve this, we model the recurrent transformations using invertible Normalizing Flows~\cite{pmlr_rezende15, dinh2017density}. 
Leveraging their invertibility and tractable Jacobians, we ensure the underlying distributions remain valid; we next detail the forward and reverse processes and the resulting training loss.

\subsection{Forward Process}
Given a sequence and text data sample pair $(x^{(j)}, u^{(j)})$, we begin by splitting each sequence into $L$ segments, denoted as $\{x^i_0\}_{i=0}^L$, each of equal length. 

For the initial segment, $x^0_0$, RDM applies Gaussian noise in a manner consistent with prior diffusion models; this process is termed \textit{diffusion-only}.
In contrast, for all subsequent segments $\{x^i_0, \; i > 0\}$, our approach diverges from previous works~\cite{pmlrv37sohldickstein15, NEURIPS2020_4c5bcfec}.
Here, RDM adds noise following the \textit{diffusion-flow} as illustrated in~\cref{fig:rfdm_arch}.
Within the \textit{diffusion-flow}, noise addition is conditioned on the previously diffused segment within the same temporal step ($x^i_{t-1}$) and the current diffusion step of the previous temporal segment ($x^{i-1}_t$).

To obtain the noisy grid samples $\{x_t^i\}_{i=0}^L$ for $t=1,..., T$, RDM first computes $x^0_t$ within the \textit{diffusion-only}, then applies normalizing flow to the corresponding temporal step.
Consistent with Ho et al.~\cite{NEURIPS2020_4c5bcfec}, the step size is controlled by the variance scheduler ${\beta_t \in (0,1)}_{t=0}^T$ for $t=0,..., T$. 
The approximate posterior is defined as:
\begin{align}
    q(x^{1:L}_{1:T}| x^0_0) &= \prod_{i=1}^L \prod_{t=1}^T q(x^i_t| x^i_{t-1}, x^{i-1}_t)
\end{align}
\subsection{Reverse Process}
The reverse process follows the 2D grid in reverse.
While the \textit{diffusion-only} reverse process uses the formulation from~\cite{pmlrv37sohldickstein15, NEURIPS2020_4c5bcfec}, for \textit{diffusion-flow} segments $\{x^i_{t-1}, i > 0\}$, the reverse process leverages temporal relations.
The generative distribution depends on the previous diffusion step ($x^i_{t}$) and the previous segment ($x^{i-1}_{t-1}$), as follows:
\begin{align} \label{eq:reverse_prob}
    p_{\theta}(x^i_{1:T}| x^i_0, x^{i-1}_T) = \prod_{t=1}^T p_{\theta}(x^i_{t-1}| x^i_t, x^{i-1}_{t-1}) \\
    p_{\theta}(x^{0:L}_{0:T}) = p(x^0_T) \prod_{i=1}^L \prod_{t=1}^T p_{\theta}(x^i_{t-1}| x^i_t, x^{i-1}_{t-1})
\end{align}

In the \textit{diffusion-only} segments, with a small $\beta_t$, the reverse distribution is Gaussian~\cite{Feller1949OnTT, pmlrv37sohldickstein15}, requiring only mean and variance estimation. 
Conversely, for \textit{diffusion-flow}, the distribution $q(x^i_t | x^i_{t-1}, x^{i-1}_t)$ is unknown because of the nonlinear flow transformations.
Consequently, the reverse distribution $p_{\theta}(x^i_{t-1} | x^i_t, x^{i-1}_{t-1})$ is also unknown, which presents a challenge for sampling.

To sample from this intractable distribution, RDM exploits the invertibility of normalizing flows~\cite{pmlr_rezende15}.
It inverts the step to \textit{diffusion-only} where distributions are Gaussian, then samples and applies the forward flow to reach the target temporal step, as described next:
\begin{align} \label{eq:gen_sample}
    x^0_{t-1} &:= \mathcal{N}(x^0_{t-1}, f_{\phi}^{(-i)}(\mu_{\theta}(x^i_t, x^{i-1}_{t-1}, t, i, c)), (1 - \Bar{\alpha}_t) \mathbf{I}) \\
    x^i_{t-1} &= f_{\phi}^{(i)}(x^0_{t-1})
\end{align}
where $\mu_{\theta}$ is a learnable neural network (denoiser); both $(-i)$ and $(i)$ indicate traversing the inverse and forward flow $i$ times, respectively; $f_{\phi}$ represents the flow model parameterized by $\phi$.
$c := \varphi(u^{(j)})$, where $\varphi$ is a pretrained CLIP model \cite{pmlr_v139_radford21a} that extracts embeddings from the textual description.

\subsection{Normalizing Flow}
We use a real-valued non-volume preserving (Real-NVP) \cite{dinh2017density} as a probabilistic transformation, as described in the following equations:
\begin{equation}
    \begin{aligned}
        &[\sigma_{\phi}, \mu_{\phi}] = f_{\phi}(x^{i-1}_t, c)) \\
        &x^i_t = x^{i-1}_t \odot \text{exp}(\sigma_{\phi}(x^{i-1}_t, c))   +\mu_{\phi}(x^{i-1}_t, c)
    \end{aligned}
\end{equation}
where $\odot$ is element-wise multiplication, and both $\sigma_{\phi}$ and $\mu_{\phi}$ are deep feed-forward networks.
When the flow $f_{\phi}$ is restricted to a linear transformation, the reverse-process distributions remain Gaussian in closed form, with the mean shifted and the variance scaled by the flow (\cref{appx:linear_flow}), reducing to a loss of the same form as Ho et al.~\cite{NEURIPS2020_4c5bcfec} with additional weighting terms. We use the more expressive non-linear coupling by default; the linear case above shows the flow need not add substantial complexity.

\subsection{Training}
During training, RDM maximizes the variational lower bound (VLB) of the log-likelihood of the 2D grid, which is simplified to (see~\cref{appx:loss_derivation} for full derivation):
\begin{equation}
    \begin{aligned}
        \mathcal{L}_{VLB} &= \underbrace{D_{kl}(q(x^{0:L}_T | x^{0:L}_0) \| p_{\theta}(x^{0:L}_T))}_{\mathcal{L}^{0:1}_T}
        + \sum_{t=2}^T \underbrace{D_{kl}(q(x^{0:L}_{t-1} | x^{0:L}_t, x^{0:L}_0) \| p_{\theta}(x^{0:L}_{t-1} | x^{0:L}_t))}_{\mathcal{L}^{0:L}_{t-1}} \\
        &\quad + \mathbb{E}_{q} [ \underbrace{ - \log (p_{\theta}(x^{0:L}_0 | x^{0:L}_1))}_{\mathcal{L}^{0:L}_0} ]
    \end{aligned}
\end{equation}

\subsubsection{KL computation}
The KL terms $\mathcal{L}^{0}_{t-1}$ within the \textit{diffusion-only} segments are straightforward, following Gaussian distributions; RDM computes them in closed form, similar to Ho \etal~\cite{NEURIPS2020_4c5bcfec}.
In contrast, the intermediate KL terms $\mathcal{L}^{1:L}_{t-1}$ pose a significant challenge because they follow unknown distributions, which is a consequence of the nonlinear flow transformations.

Learning the posterior of unknown distributions often requires methods like Markov Chain Monte Carlo (MCMC)~\cite{Metropolis1953, Vono20}, but with $1000$ diffusion steps, these methods are infeasible due to their slow convergence. 
This raises a crucial question: how can we learn the reverse process efficiently?

To train efficiently, RDM uses the invertibility of normalizing flows (\cref{back:flow}) to derive a closed-form loss. 
It employs an inverse flow to map predicted samples from temporal step $i$ back to temporal step $0$ (\textit{diffusion-only}), where the distributions are Gaussian, and then computes the KL in closed form.
The result is then transformed back to temporal step $i$ via the forward flow.

Consistent with Ho \etal~\cite{NEURIPS2020_4c5bcfec}, we learn only the mean of the distribution and fix the variance to $\sigma_t = \beta_t$.
This approach yields a formula similar to that of~\cite{NEURIPS2020_4c5bcfec}:

\begin{equation} \label{eq:simple_loss}
    \begin{aligned}
        &\mathcal{L}_{t-1}^{i} = 
        \mathbb{E}_{q} \left[w(t) \|x^i_0 - f_{\theta}(x^i_t, x^{i-1}_{t-1},t,i)\|^2\right]
    \end{aligned}
\end{equation}
Here, $x^i_0$ is the clean segment at temporal step $i$ and $f_{\theta}$ is a learnable neural network that predicts a clean segment; refer to~\cref{appx:loss_derivation} for a full derivation.

\subsubsection{Training protocol}
We optimize the parameters of both the flow ($\phi$) and the denoiser ($\theta$) jointly by minimizing the expectation in \cref{eq:simple_loss}. 
This expectation is approximated via Monte Carlo integration, where we uniformly sample a diffusion step $t$ and a temporal step $i$ at each iteration.
The training sequence is then processed by traversing the normalizing flow to reach the sampled temporal position within the dependency grid (see~\cref{fig:rfdm_arch}).
The complete training sequence and the explicit diffusion update mechanics are detailed in Algorithm \ref{alg:diffusion_update}.

\subsubsection{Sampling} \label{sec:diff_sample}
Unlike existing autoregressive frameworks~\cite{ZhaoDart2025, shi2024amdm, meng2024rethinking, tevet2025closd, han2023amd}, RDM does not require fully denoising preceding segments before denoising subsequent ones. 
Instead, leveraging the flow interface, RDM skips intermediate diffusion steps by traversing a diagonal ``staircase" pattern across the 2D dependency grid (\cref{fig:rfdm_arch}). 
Empirically, we find that setting the initial step of the staircase equal to the total number of segments yields an optimal balance between computational complexity and synthesis quality (see \cref{appx:rfdm_properties,appx:diff_sample} for detailed analyses). 
The complete inference procedure is formalized in Algorithm \ref{alg:rfdm_sample_staircase}.

\noindent
\begin{minipage}[t]{0.48\textwidth}
\begin{algorithm}[H]
\footnotesize
\caption{\texttt{diffusion\_update\_step}}\label{alg:diffusion_update}
\begin{algorithmic}[1]
\State \textbf{repeat}
\IndState[0.1] $x_0 \sim q(\mathcal{X})$ {\small // sample sequence}
\IndState[0.1] $x^0_0:= select\_at([x^i]_{i=0}^L, 0)$
\IndState[0.1] $t \sim Uniform({1,...,T})$
\IndState[0.1] $i \sim Uniform({0,...,L})$
\IndState[0.1] $\epsilon^0_0 \sim \mathcal{N}(\mathbf{0}, \mathbf{I})$
\IndState[0.1] $x^i_t = f^{(i)}_{\phi}(\sqrt{\bar{\alpha}_t} x^0_0 + \sqrt{1 - \bar{\alpha}_t} \epsilon^0_t)$
\IndState[0.1] $x^{i-1}_t = f^{(i-1)}_{\phi}(\sqrt{\bar{\alpha}_{t-1}} x^0_0 + \sqrt{1 - \bar{\alpha}_{t-1}} \epsilon^0_{t-1})$
\IndState[0.1] Take gradient step on
\IndStatex[0.3] $\nabla_{\theta} \| x^i_0 - f_{\theta}(x^i_t, x^{i-1}_{t-1},t,i) \|^2$
\State \textbf{until} end of epoch
\end{algorithmic}
\end{algorithm}
\end{minipage}
\hfill
\begin{minipage}[t]{0.48\textwidth}
\begin{algorithm}[H]
\footnotesize
\caption{\texttt{Sampling}}\label{alg:rfdm_sample_staircase}
\begin{algorithmic}[1]
\State $x^0_T \sim \mathcal{N}(0, I)$, $\mathcal{S}_{out} := [x^0_T, f_{\phi}(x^0_T)]$
\State \textbf{for} $t = T, ..., 1$ \textbf{do}
\IndState[0.1] $\mathcal{S} = \mathcal{S}_{out}$
\IndState[0.1] \textbf{for} $j = 0, ..., i+1$ \textbf{do}
\IndState[0.2] $z \sim \mathcal{N}(0, I)$ if $t > 1$ else $z = \mathbf{0}$
\IndState[0.2] $x^j_t = S[j]$
\IndState[0.2] $x^{j-1}_{t-1} = S_{out}[j-1]$ if $j > 0$ else $x^{j-1}_{t-1} = \mathbf{0}$
\IndState[0.2] $x^0_t = f^{(-j)}_{\phi}(S[j])$
\IndState[0.2] $x^0_{t - 1} = f^{(-j)}_{\phi}(f_{\theta}(x^j_t, x^{j-1}_{t-1}, t, j))$
\IndState[0.2] $x^j_{t-1} = f^{(j)}_{\phi}(x^0_{t-1})$
\IndState[0.2] $\mathcal{S}_{out}.append(x^j_{t-1})$
\IndState[0.1] \textbf{end for}
\State \textbf{end for}
\State \textbf{return} $\mathcal{S}_{out}$
\end{algorithmic}
\end{algorithm}
\end{minipage}

\begin{table*}[t]
\resizebox{\textwidth}{!}{
\begin{tabular}{lllllllll}
\hline
\multicolumn{1}{c}{\multirow{2}{*}{Framework}} & \multirow{2}{*}{Method} & \multicolumn{3}{c}{R-Precision ($\uparrow$)} & \multirow{2}{*}{FID ($\downarrow$)} & \multirow{2}{*}{MultiModal Dist ($\downarrow$)} & \multirow{2}{*}{MultiModality ($\uparrow$)} \\ \cline{3-5}
\multicolumn{1}{c}{}                        &                   & Top 1     & Top 2    & Top 3    &                      &                                  &                            &                                \\ \hline
                                            & Real motion       & $0.511 ^{\pm.003}$ & $0.703 ^{\pm.002}$ & $0.797 ^{\pm.003}$ & $0.002 ^{\pm.000}$ & $2.974 ^{\pm.008}$ & - \\ \hline
\multirow{3}{*}{VAE}                        & TEMOS~\cite{TEMOS_Petrovich2022}             & $0.424 ^{\pm.002}$ & $0.612 ^{\pm.002}$ & $0.722 ^{\pm.002}$ & $3.734 ^{\pm.028}$ & $3.703 ^{\pm.008}$ & $0.368 ^{\pm.018}$ \\
                                            & T2M~\cite{t2m_Guo_2022}               & $0.457 ^{\pm.002}$ & $0.639 ^{\pm.003}$ & $0.740 ^{\pm.003}$ & $1.067 ^{\pm.002}$ & $3.340 ^{\pm.008}$ & $2.090 ^{\pm.083}$ \\
                                            & MoMask~\cite{guo2024momask}            & $0.521^{\pm.002}$  & $0.713^{\pm.002}$  & $0.807^{\pm.002}$  & $0.045^{\pm.002}$  & $2.958^{\pm.008}$  & $1.241^{\pm.040}$ \\ \hline

\multirow{8}{*}{Volume Diffusion}           & MDM~\cite{tevet2023human}               & $0.320^{\pm.005}$     & $0.498^{\pm.004}$     & $0.611^{\pm.007}$     & $0.544^{\pm.044}$     & $5.566^{\pm.027}$     & \bstcell$2.799^{\pm.072}$ \\
                                            & MotionDiffuse~\cite{zhang2022motiondiffuse}     & $0.491^{\pm.001}$     & $0.681^{\pm.001}$     & $0.782^{\pm.001}$     & $0.630^{\pm.001}$     & $3.113^{\pm.001}$     & $1.553^{\pm.042}$ \\
                                            & MLD~\cite{chen2023executing}               & $0.481^{\pm.003}$     & $0.673^{\pm.003}$     & $0.772^{\pm.002}$     & $0.473^{\pm.013}$     & $3.196^{\pm.010}$     & \midcell$2.413^{\pm.079}$ \\
                                            & MotionMamba~\cite{zhang2025motion}       & \worcell$0.502^{\pm.003}$     & \worcell$0.693^{\pm.002}$     & \worcell$0.792^{\pm.002}$     & \worcell$0.281^{\pm.009}$     & \worcell$3.060^{\pm.058}$     & \worcell$2.294^{\pm.058}$ \\
                                            & ReMoDiffuse~\cite{Zhang_2023_ICCV}       & \midcell$0.510^{\pm.005}$     & \midcell$0.698^{\pm.006}$     & \midcell$0.795^{\pm.004}$     & \midcell$0.103^{\pm.004}$     & \bstcell$2.974^{\pm.016}$     & $1.795^{\pm.043}$ \\   
                                            & Light-T2M~\cite{light_t2m_2025}         & \bstcell$0.511^{\pm.003}$     & \bstcell$0.699^{\pm.002}$     & \bstcell$0.795^{\pm.002}$     & \bstcell$0.040^{\pm.002}$     & \midcell$3.002^{\pm.008}$     & $1.670^{\pm.061}$ \\    
                                            & \shortmethod-196 (ours)    & $0.449^{\pm.001}$     & $0.661^{\pm.001}$     & $0.777^{\pm.001}$     & $0.585^{\pm.001}$     & $3.113^{\pm.001}$     & $1.538^{\pm.042}$ \\ \hline

\multirow{7}{*}{Segment-based diffusion}          & AMD~\cite{han2023amd}               & $-$     & $-$     & $0.617^{\pm.014}$ & $0.586 ^{\pm.107}$ & $5.469 ^{\pm.063}$ & \bstcell$2.512 ^{\pm.232}$ \\
                                            & CLoSD (DiP)~\cite{tevet2025closd}   & \bstcell$0.463^{\pm.006}$ & \bstcell$0.668^{\pm.006}$ & \bstcell$0.773^{\pm.008}$           & \bstcell$0.247^{\pm.034}$ & \bstcell$3.178^{\pm.025}$ & $-$ \\
                                            & A-MDM~\cite{shi2024amdm}             & $-$     & $-$     & $-$               & $1.7435^{\pm.0813}$ & $-$ & $-$ \\
                                            & MD-49      & $0.339^{\pm.005}$ & $0.499^{\pm.006}$ & $0.605^{\pm.007}$ & $4.414^{\pm.116}$ & $4.338^{\pm.031}$ & \midcell$2.262^{\pm.155}$ \\
                                            & MD-28      & $0.368^{\pm.002}$ & $0.546^{\pm.007}$ & $0.658^{\pm.007}$ & $3.377^{\pm.0}$ & \worcell$4.019^{\pm.03}$ & \worcell$1.771^{\pm.145}$ \\
                                            & \shortmethod-49 (ours)   & \worcell$0.442^{\pm.006}$ & \midcell$0.664^{\pm.004}$ & \worcell$0.762^{\pm.004}$ & \worcell$0.312^{\pm.016}$ & \midcell$3.604^{\pm.006}$ & $1.544^{\pm.046}$ \\
                                            & \shortmethod-28 (ours)   & \midcell$0.455^{\pm.002}$ & \worcell$0.653^{\pm.003}$ & \midcell$0.772^{\pm.002}$ & \midcell$0.272^{\pm.018}$ & \worcell$3.651^{\pm.008}$ & $1.199^{\pm.04}$ \\ \hline
\end{tabular}
}
\caption{
\textbf{Quantitative results on the HumanML3D~\cite{Guo_2022_CVPR} test set.}
All methods use ground truth motion length.
Evaluations were run 20 times with $95\%$ CI. 
Rows are colour-coded as \bsttxt{best}, \midtxt{second best}, and \wortxt{third best}.
Our recurrent methods (\shortmethod-49/28) outperform segment-based baselines significantly and are comparable to CLoSD (DiP)~\cite{tevet2025closd}.
}
\label{tab:hml_metrics}
\end{table*}

\begin{table*}[t]
\resizebox{\textwidth}{!}{\begin{tabular}{lllllllll}
\hline
\multicolumn{1}{c}{ \multirow{2}{*}{Framework}} & \multirow{2}{*}{Method} & \multicolumn{3}{c}{R-Precision ($\uparrow$)} & \multirow{2}{*}{FID ($\downarrow$)} & \multirow{2}{*}{MultiModal Dist ($\downarrow$)} & \multirow{2}{*}{MultiModality ($\uparrow$)} \\ \cline{3-5}
\multicolumn{1}{c}{}                        & & Top 1     & Top 2    & Top 3    &                      &                                  &                            &                                \\ \hline
                                            & Real motion     & $0.5138$            & $0.6957$              & $0.7937$             & $0.002^{\pm.000}$     & $2.974^{\pm.008}$     & - \\ \hline
\multirow{3}{*}{VAE}                        & TEMOS~\cite{TEMOS_Petrovich2022}           & $0.353^{\pm.006}$   & $0.561^{\pm.007}$     & $0.687^{\pm.005}$    & $3.717^{\pm.051}$     & $3.417^{\pm.019}$     & $0.532^{\pm.034}$ \\
                                            & T2M~\cite{t2m_Guo_2022}             & $0.370^{\pm.005}$   & $0.569^{\pm.007}$     & $0.693^{\pm.007}$    & $2.770^{\pm.109}$     & $3.401^{\pm.008}$     & $1.482^{\pm.065}$ \\
                                            & MoMask~\cite{guo2024momask}            & $0.433^{\pm.007}$     & $0.656^{\pm.005}$     & $0.781^{\pm.005}$     & $0.204^{\pm.011}$     & $2.779^{\pm.022}$     & $1.131^{\pm.043}$ \\ \hline

\multirow{8}{*}{Volume Diffusion}           & MDM~\cite{tevet2023human}             & $0.164^{\pm.004}$   & $0.291^{\pm.004}$     & $0.396 ^{\pm.004}$    & $0.497^{\pm.021}$     & $9.191^{\pm.022}$     &\midcell$1.907^{\pm.214}$ \\
                                            & MotionDiffuse~\cite{zhang2022motiondiffuse}   & $0.417^{\pm.004}$   & $0.621^{\pm.004}$     & \worcell$0.739^{\pm.004}$     & $1.954^{\pm.062}$     & \worcell$2.958^{\pm.005}$     & $0.730^{\pm.013}$ \\
                                            & MLD~\cite{chen2023executing}             & $0.390^{\pm.008}$   & $0.609^{\pm.008}$     & $0.734^{\pm.007}$     & $0.404^{\pm.027}$     & $3.204^{\pm.027}$     &\bstcell$2.192^{\pm.071}$ \\
                                            & MotionMamba~\cite{zhang2025motion}     & \worcell$0.419^{\pm.006}$   & \midcell$0.645^{\pm.005}$     & \midcell$0.765^{\pm.006}$     &\worcell$0.307^{\pm.041}$     &$3.021^{\pm.025}$     &\worcell$1.678^{\pm.064}$ \\
                                            & ReMoDiffuse~\cite{Zhang_2023_ICCV}     & \midcell$0.427^{\pm.014}$     & \worcell$0.641^{\pm.004}$     & \midcell$0.765^{\pm.055}$     & \bstcell$0.155^{\pm.006}$     & \midcell$2.814^{\pm.012}$     & $1.239^{\pm.028}$ \\
                                            & Light-T2M~\cite{light_t2m_2025}       & \bstcell$0.444^{\pm.006}$     & \bstcell$0.670^{\pm.007}$     & \bstcell$0.794^{\pm.005}$     & \midcell$0.161^{\pm.009}$     & \bstcell$2.746^{\pm.016}$     & $1.005^{\pm.036}$ \\
                                            & \shortmethod-196 (ours)  & $0.381^{\pm.004}$   & $0.603^{\pm.004}$    &$0.734^{\pm.004}$    & $1.8151^{\pm.062}$   & \worcell$2.958^{\pm.005}$     & $0.693^{\pm.013}$ \\ \hline

\multirow{4}{*}{Segment-based diffusion}          & MD-49   & \worcell$0.3067^{\pm.007}$ & \worcell$0.4816^{\pm.007}$ & \worcell$0.6085^{\pm.007}$ & \worcell$2.8553^{\pm.073}$ & \worcell$4.1278^{\pm.026}$ & \bstcell$2.9595^{\pm.086}$ \\
                                            & MD-28   & $0.2728^{\pm.004}$ & $0.4528^{\pm.0059}$ & $0.5794^{\pm.005}$ & $5.5562^{\pm.085}$ & $4.5790^{\pm.03}$ & \midcell$2.4437^{\pm.1054}$ \\
                                            & \shortmethod-49 (ours) & \midcell$0.364^{\pm.008}$ & \midcell$0.576^{\pm.008}$ & \midcell$0.712^{\pm.006}$ & \midcell$1.95^{\pm.051}$ & \midcell$3.541^{\pm.02}$ & \worcell$1.068^{\pm.045}$ \\
                                            & \shortmethod-28 (ours) & \bstcell$0.377^{\pm.005}$ & \bstcell$0.5913^{\pm.003}$ & \bstcell$0.715^{\pm.005}$ & \bstcell$1.819^{\pm.05}$ & \bstcell$3.412^{\pm.016}$ & $0.59^{\pm.038}$ \\ \hline
\end{tabular}}
\caption{
\textbf{Quantitative results on the KIT-ML~\cite{Plappert2016} test set.}
All methods use ground truth motion length.
Evaluations were run 20 times with $95\%$ CI. 
Rows are colour-coded as \bsttxt{best}, \midtxt{second best}, and \wortxt{third best}.
Our recurrent methods (\shortmethod-49/28) significantly outperform segment-based baselines.
MD-x baselines degrade with more segments, because the dataset features shorter sequences (max $196$ frames), which harms future segment prediction.
Volume diffusion models generally outperform segment-based methods.
}
\label{tab:kit_metrics}
\end{table*}

\begin{table*}[t]
\resizebox{\textwidth}{!}{
\begin{tabular}{llcllllllcllllllc}
\hline
\multicolumn{1}{c}{\multirow{3}{*}{Framework}} 
    &
  \multicolumn{1}{c}{\multirow{3}{*}{Method}} 
    &
  \multicolumn{6}{c}{Total Inference Time (s) $\downarrow$} &
   &
   &
  \multicolumn{5}{c}{FLOPs (T) $\downarrow$} &
   &
  \multirow{3}{*}{Parameters} \\ \cline{3-8} \cline{10-15}
\multicolumn{1}{c}{} & &
  \multicolumn{3}{c}{DDIM} &
   &
   &
  DDPM &
   &
  \multicolumn{3}{c}{DDIM} &
   &
   &
  DDPM &
   &
   \\ \cline{3-6} \cline{8-8} \cline{10-13} \cline{15-15}
\multicolumn{1}{c}{} & &
  \multicolumn{1}{l}{10} &
  50 &
  100 &
  200 &
   &
  1000 &
   &
  \multicolumn{1}{l}{10} &
  50 &
  100 &
  200 &
   &
  1000 &
   &
   \\ \hline
\multirow{1}{*}{Volume Diffusion}   & MotionDiffuse~\cite{zhang2022motiondiffuse}    & $-$ & \worcell$2.71^{\pm0.23}$ & \worcell$4.92^{\pm.04}$  & \worcell$9.79^{\pm.08}$  &  & \worcell$52.07^{\pm2.06}$   &  & $-$ & \worcell$64.64$ & \worcell$128.44$ & \worcell$256.02$ &  & \worcell$1276.71$ &  & $x \in \mathbb{R}^{196 \times 251}$ \\ \hline
\multirow{5}{*}{Segment-based diffusion}  & CLoSD (DiP)~\cite{tevet2025closd}                    & \worcell$1.77^{\pm.037}$ & $8.73^{\pm.03}$ & $17.5^{\pm.09}$    & $34.87{\pm.04}$ &  & $(93.18^{\pm.16})$   &  & \worcell$45.89$ & $229.44$ & $458.88$  & $917.76$  &  & $(2617.93)$    &  & $x \in \mathbb{R}^{40 \times 251}$  \\
                                    & MD-28  & $-$ & $5.40^{\pm.08}$ & $11.39^{\pm.80}$    & $21.34{\pm.09}$ &  & $113.20^{\pm4.63}$   &  & $-$ & $83.41$ & $165.97$  & $331.08$  &  & $1652.01$    &  & $x \in \mathbb{R}^{28 \times 251}$  \\
                                    & \shortmethod-49 (ours)  & \bstcell$0.21^{\pm.007}$ & \bstcell$1.95^{\pm1.71}$ & \bstcell$2^{\pm1.19}$    & \bstcell$3.26^{\pm1.51}$ &  & \bstcell$10.89^{\pm2.11}$   &  & \bstcell$0.275$ & \bstcell$0.282$ & \bstcell$0.458$  & \bstcell$0.809$  &  & \midcell$3.62$    &  & $x \in \mathbb{R}^{49 \times 251}$  \\
                                    & \shortmethod-28 (ours)  & \midcell$0.59^{\pm.005}$ & \midcell$2.61^{\pm1.75}$ & \midcell$2.48^{\pm1.19}$ & \midcell$3.86{\pm1.49}$  &  & \midcell$11.7^{\pm2.02}$    &  & \midcell$0.66$ & \midcell$0.533$ & \midcell$0.644$  & \midcell$0.866$  &  & \bstcell$2.32$    &  & $x \in \mathbb{R}^{28 \times 251}$  \\ \hline
\end{tabular}}
\caption{
\textbf{Inference time comparison.}
Rows are colour-coded as \bsttxt{best}, \midtxt{second best}, and \wortxt{third best}.
Our method skips diffusion steps via \textit{staircase sampling}, reducing inference time and FLOPs compared to volume methods that process the entire sequence. 
RDM is more efficient than segment-based baselines (MD-x/CLoSD~\cite{tevet2025closd}) because it does not fully denoise previous frames to generate new ones.
}
\label{tab:compute_cost}
\end{table*}

\begin{figure*}[t]
     \centering
     \begin{minipage}[b]{0.48\textwidth}
         \centering
         \includegraphics[width=\textwidth]{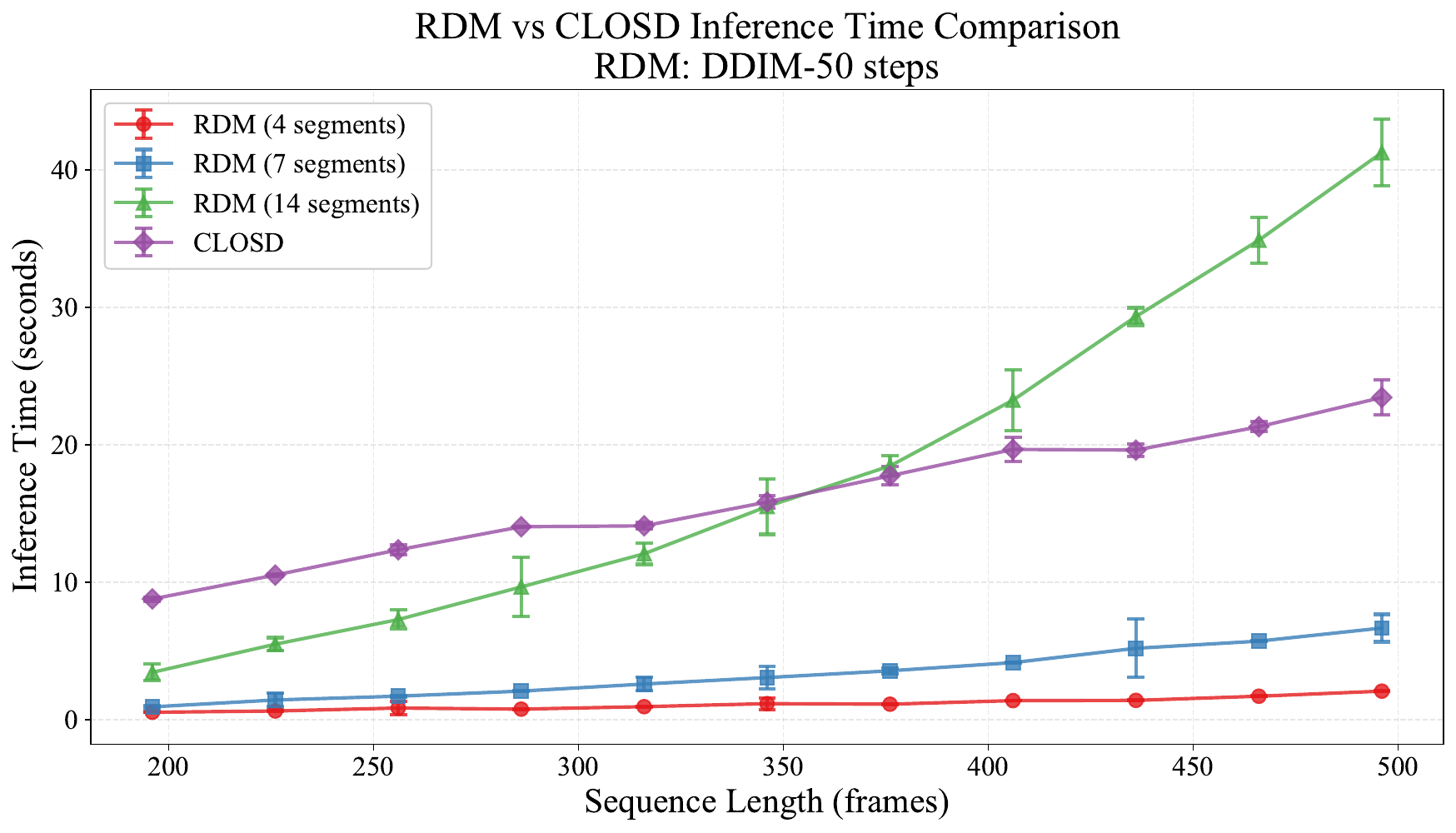}
         \label{fig:image_left}
     \end{minipage}
     \hfill
     \begin{minipage}[b]{0.48\textwidth}
         \centering
         \includegraphics[width=\textwidth]{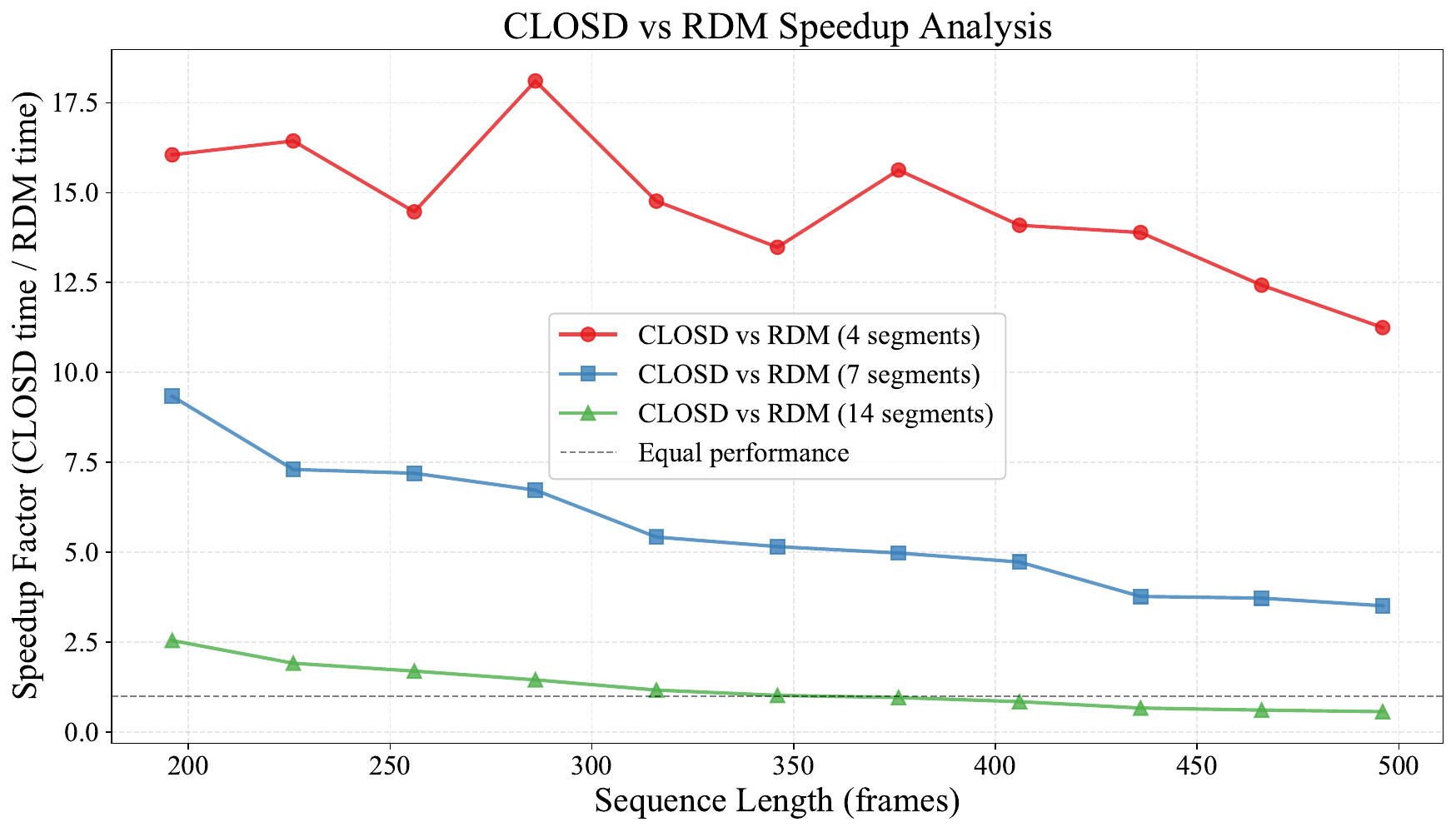}
         \label{fig:image_right}
     \end{minipage}
     
     \caption{
     \textbf{Computational efficiency comparison between \shortmethod and CLoSD (DiP)~\cite{tevet2025closd}.}
     (left) Inference time vs. sequence length for \shortmethod with 4, 7, 14 segments (DDIM-50 steps) and CLoSD. 
     Error bars show $\pm3$ standard deviations. (right) Speedup analysis dividing CLoSD inference time by \shortmethod's. 
     \shortmethod-49/28 outperforms CLoSD by $(3.51 \text{ to } 18.11)\times$ across sequence lengths, with higher speedups for shorter sequences. 
     Speedup varies with segment configuration: \shortmethod-49 is faster than \shortmethod-28, \shortmethod-14 is slowest.
     }
     \label{fig:rollout_compute_comparision}
\end{figure*}

\section{Experiments} \label{sec:exp}
The primary objective of our evaluation is to quantify the performance and stability of the recurrent diffusion framework across distinct generative tasks. 
Unlike volume-diffusion methods where the computational cost scales prohibitively with sequence length, segment-based generation must instead manage the potential accumulation of temporal errors across sequential segments.
For a controlled comparison across segment configurations, we denoise in pose space rather than latent space, following MotionDiffuse~\cite{zhang2022motiondiffuse} and Light-T2M~\cite{light_t2m_2025}: pose-space frames skip the pretrained VAE, so segment length and count are simply chunking parameters we can vary freely to ablate their impact on generative stability, without retraining a latent encoder per configuration.

\subsection{Evaluation Tasks and Baselines}
We evaluate our framework on two tasks: standard text-to-motion generation and extended sequence synthesis (generating motion beyond the training horizon), comparing against both volume-diffusion and segment-based diffusion baselines as defined in \cref{sec:intro,sec:rw}.

We benchmark RDM on text-to-motion generation against volume-diffusion models — MDM~\cite{tevet2023human}, MotionDiffuse~\cite{zhang2022motiondiffuse}, MLD~\cite{chen2023executing}, MotionMamba~\cite{zhang2025motion}, Light-T2M~\cite{light_t2m_2025} — non-diffusion approaches — TEMOS~\cite{TEMOS_Petrovich2022}, T2M~\cite{t2m_Guo_2022}, MoMask~\cite{guo2024momask} — and, on fixed-length sequences, segment-based autoregressive baselines AMD~\cite{han2023amd}, CLoSD~\cite{tevet2025closd}, and A-MDM~\cite{shi2024amdm}.

For extended sequence synthesis, we compare against segment-based models: CLoSD~\cite{tevet2025closd} and our MotionDiffuse-backed MD-49 and MD-28 baselines (named by per-segment frame length over the $196$-frame training horizon, i.e., $4$ and $7$ segments, respectively; we use the same convention for \shortmethod-$L$).
These MD-$x$ variants serve as non-recurrent ablations where segments are denoised independently.
Other segment-based methods, such as TEDi~\cite{zhang2024tedi} and UniPhys~\cite{wu2025uniphys}, are omitted due to dataset discrepancies.

\subsection{Experimental setup}
\noindent{\textbf{Datasets.}}
The KIT-ML dataset~\cite{Plappert2016} provides $3911$ motion sequences and $6353$ sequence-level natural language descriptions. 
HumanML3D~\cite{Guo_2022_CVPR} re-annotates the AMASS dataset~\cite{AMASS_ICCV_2019} and the HumanAct12 dataset~\cite{guo2020action2motion}. 
It provides $44970$ annotations on $14616$ sequences.

\noindent{\textbf{Evaluation Metrics.}}
We adhere to evaluation protocols from~\cite{zhang2022motiondiffuse, chen2023executing, zhang2025motion, light_t2m_2025}, assessing methods with four metrics (R-Precision, Fréchet Inception Distance (FID), Multi-modal Distance and Multimodality), each repeated $20$ times with a $95\%$ confidence interval.
\textbf{R-Precision}: measures motion-to-text retrieval accuracy by randomly selecting 31 mismatched text prompts and reporting top-k accuracies based on embeddings. 
\textbf{Fréchet Inception Distance (FID)}: evaluates realism and diversity of motions by comparing distributions of embeddings from generated and ground-truth data via a pretrained motion encoder. 
\textbf{Multi-modal Distance}: uses a pretrained model to embed motions and calculates Euclidean distance between text and motion features. 
\textbf{Multimodality}: assesses the diversity of motions from a single text by measuring joint position differences across $32$ generated sequences.
\subsection{Qualitative evaluation}
\begin{figure}[t]
    \centering
    \includegraphics[width=1.0\linewidth]{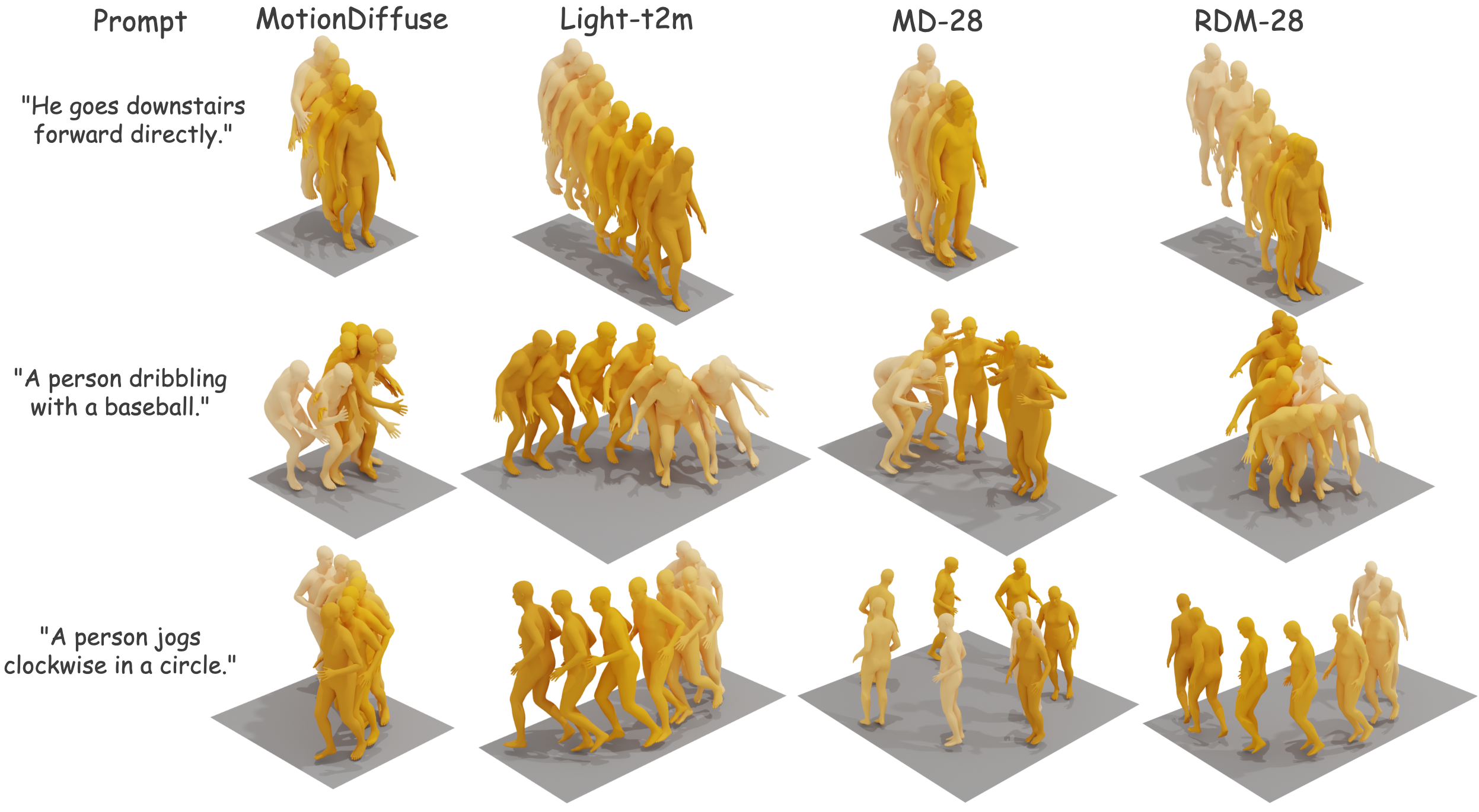}
    \caption{
    \textbf{Qualitative comparison on HumanML3D~\cite{Guo_2022_CVPR}.}
    Additional stacked-SMPL qualitative examples comparing volume-diffusion methods --- MotionDiffuse~\cite{zhang2022motiondiffuse}, and Light-T2M~\cite{light_t2m_2025} --- against segment-based methods --- MD-28 and \shortmethod-28 --- showing all five produce motions aligned with the input prompts.
    }
    \label{fig:smpl_mstack_main_fig}
\end{figure}
\cref{fig:smpl_mstack_main_fig,fig:skeleton_figs} show our methods produce plausible motions from input captions.
To evaluate extended generation, we prompt for continuous, meaningful motion beyond the horizon using open-ended descriptions and visually inspect for failures.

In \cref{fig:skeleton_figs}, with the prompt \textit{``A person is dribbling with a basketball"}, we extend motion beyond the training horizon to compare MD-49, MD-28, and \shortmethod-28. 
Green circles mark frames where the dribbling motion remains correct and red circles mark frames where it breaks down: within the training horizon (left of the dashed line), MD-49 dribbles correctly at first but loses the motion near the horizon boundary, a problem not observed in MD-28 or \shortmethod-28.
Beyond the training horizon (right of the dashed line), MD-49 fails to produce plausible motion, while MD-28 and \shortmethod-28 generate sequences that reasonably match the input prompt overall; however, MD-28's hands also stop performing the dribbling motion once past the horizon, whereas \shortmethod-28 keeps dribbling coherently throughout.
Nevertheless, we observe that \shortmethod-28’s motions are more coherent than MD-28, and significantly faster (see~\cref{tab:compute_cost}).
Static images do not accurately convey motion quality; we recommend that the reader view our supplementary video.
\subsection{Quantitative evaluation}
Analyzing segment-based methods in ~\cref{tab:hml_metrics} and \ref{tab:kit_metrics} shows MD-49/28 are strong baselines with $0.658$ R-Top3, comparable to AMD~\cite{han2023amd}. 
Our \shortmethod-49/28 outperform MD-x and other baselines except CLoSD (DiP)~\cite{tevet2025closd} on HumanML3D~\cite{Guo_2022_CVPR} and KIT-ML~\cite{Plappert2016}, highlighting the importance of a recurrent connection.
MD-x baselines also degrade with more segments on the shorter-sequence KIT-ML dataset but not on HumanML3D, whose longer sequences ease future-segment prediction (see \cref{tab:kit_metrics}).
Our \shortmethod-28 achieves comparable Top-3 R-precision to CLoSD (DiP) ($0.772$ vs.\ $0.773$), while CLoSD (DiP) attains a better FID ($0.247$ vs.\ $0.272$); overall, \shortmethod remains competitive on motion fidelity while offering substantially lower inference cost (see \cref{tab:compute_cost}).

As evident in \cref{tab:hml_metrics} and~\ref{tab:kit_metrics}, volume diffusion outperforms segment-based methods, with \shortmethod-196 achieving results comparable to MotionDiffuse~\cite{zhang2022motiondiffuse} on KIT-ML and HumanML3D.
Their generation is limited to the training horizon.

\subsection{Inference time}
\shortmethod accelerates inference by skipping diffusion steps, maintaining compatibility with DDIM sampling~\cite{song2021denoising}. 
As shown in \cref{tab:compute_cost}, \shortmethod variants achieve lower FLOPs and significantly reduced wall-clock times compared to MotionDiffuse~\cite{zhang2022motiondiffuse} and CLoSD~\cite{tevet2025closd}. 
Using DDIM-50 to balance fidelity and speed, we evaluate scaling properties in \cref{fig:rollout_compute_comparision}. 
\shortmethod-49 is the most efficient configuration, delivering an $11.25\times$ to $18.11\times$ speedup over CLoSD, followed by \shortmethod-28 at $3.51\times$ to $9.34\times$. 

Efficiency is inherently tied to segment configuration and sequence length; while most variants scale favorably, \shortmethod-14’s overhead leads to diminished returns beyond the 350-frame threshold (see \cref{appx:rfdm_properties,appx:diff_sample}). These benchmarks were conducted on an NVIDIA H100 GPU using KIT-ML data, with results averaged over $20$ trials.
Detailed comparative analyses are available in~\cref{appx:diff_sample}.

\begin{wrapfigure}{r}{0.45\textwidth}
    \centering
    \includegraphics[width=0.42\textwidth]{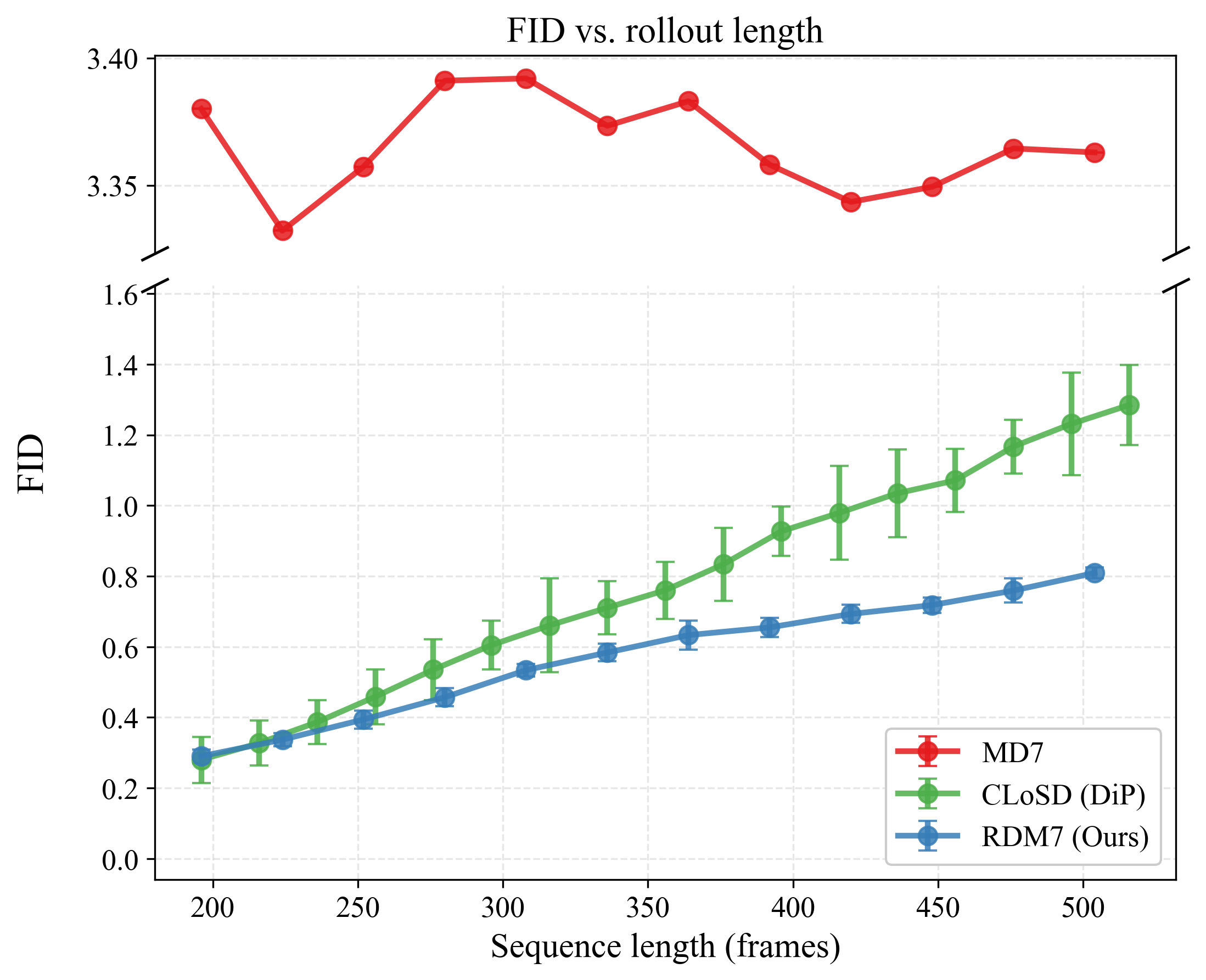}
    \caption{
    \textbf{FID stability beyond the training horizon} (lower is better; note the broken $y$-axis). \shortmethod-28 stays consistently below CLoSD (DiP) and MD-28 as generation extends well past the $196$-frame training horizon.
    }
    \label{fig:fid_extended_egn}
\end{wrapfigure}

\noindent{\textbf{Beyond the training horizon evaluation.}}
We quantitatively evaluate long-sequence generation beyond the training horizon (see~\cref{fig:fid_extended_egn}). We extended inference past $L=196$ frames and generated sequences of up to 500 frames, corresponding to approximately 10 segments.
\needspace{9cm}
We compare \shortmethod with the autoregressive baselines MD-28 and CLoSD (DiP), computing FID between the generated long-horizon sequences and the ground-truth motion distribution over the training horizon.
MD-28’s FID remains approximately constant beyond the training horizon.
This is because MD-28 assigns each segment a positional index learned during training; when generation extends beyond the supported horizon, these indices repeat cyclically, causing the model to reproduce previously learned motion rather than generate new long-horizon dynamics.
\shortmethod-28 maintains lower FID than CLoSD over long-horizon, while not requiring ground-truth motion as input.
CLoSD uses a segment size of 20 frames, different from MD-28 and \shortmethod-28, explaining the slight misalignment in \cref{fig:fid_extended_egn}.
At approximately $L=500$ frames ($L=516$ for CLoSD and $L=504$ for \shortmethod-28), \shortmethod-28 achieves an FID that is $\approx 1.6\times$
lower than CLoSD and $\approx 6\times$ faster.
\section{Conclusion}
\label{sec:con}
\noindent We introduce Recurrent Diffusion Model (\shortmethod), a temporal diffusion framework that enforces temporal constraints by recurrently adding and removing noise.
\shortmethod{}’s volume method matches SOTA volume results. While the recurrent version is on par with autoregressive diffusion, it generates sequences beyond the training horizon with significantly lower inference complexity. We believe \shortmethod advances temporal diffusion models and naturally extends into the latent space, though these avenues remain for future research.\\

\noindent{\textbf{Limitations.}} \label{sec:limitation}
Although \shortmethod can generate sequences beyond the training horizon, its numerical stability depends on the flow, which is challenging as the number of segments increases.
Additionally, generated frames become noisier over time, especially with a few recurrence steps. (See \cref{appx:nf_nistability})

\section*{Acknowledgements}
The research presented here has been supported by the UCL Centre for Doctoral Training in Foundational AI under UKRI grant number EP/S021566/1 and the UKRI/EPSRC AI Hub in Generative Models under grant number EP/Y028805/1.
RB acknowledges support from the EPSRC (EP/V026259/1).
The authors are grateful to the Baskerville Tier 2 HPC service (EP/T022221/1, EP/W032244/1). 
We thank Shalini Maiti, Abdallah Basheir, Wonbong Jang, Oscar Key, and Waleed Dawood for fruitful discussions and valuable feedback.
We also thank Third Dimension, the startup where the lead author works, for their continued support and encouragement.

\bibliography{main}

\clearpage
\appendix
\setcounter{page}{1}

\begin{center}
    \LARGE\bfseries\sffamily\textcolor{bmv@sectioncolor}{Supplementary Material}
\end{center}
\vspace{1em}

\begin{figure}[t]
    \centering
    \includegraphics[width=0.4\textwidth]{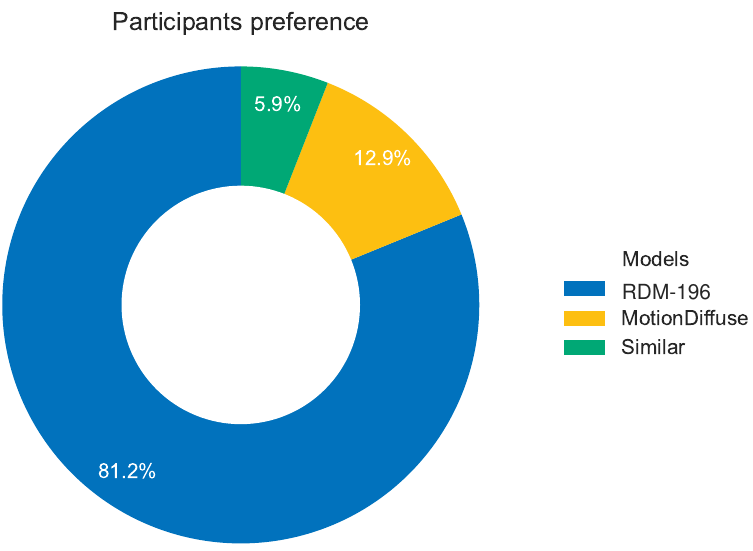}
    \hspace{1em}
    \includegraphics[width=0.4\textwidth]{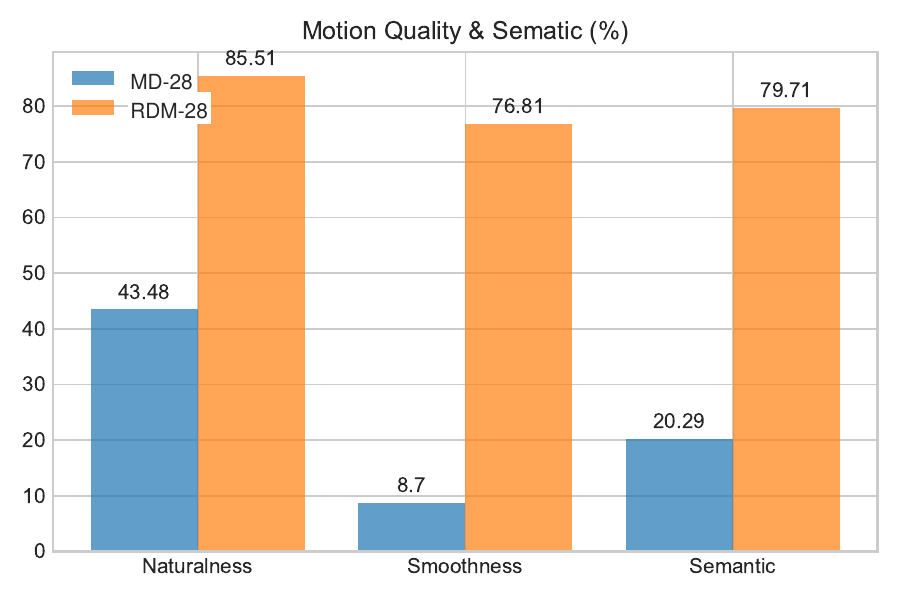}
    \caption{
    \textbf{User study analysis results.}
    Semantic quality in percentage denotes how well the motion visually corresponds to the input sentences.
    Motion quality in percentage shows the overall quality of motion in terms of naturalness.
    Smoothness in percentage shows how good the overall motion smoothness and coherence are.
    (A higher value is better.)
    Our method’s score, denoted in \shortmethod-28, has the highest percentage compared to the baseline.
    \shortmethod-196 receives the largest share of participant preference (left).}
    \label{fig:user_study}
\end{figure}

\noindent\textbf{A Note on Terminology in the Supplementary Video:} Please note that in the accompanying supplementary video, the framework's overarching category is referred to as \textit{``rollout diffusion''}. To establish a more rigorous and standard taxonomy, this term was updated to \textit{``segment-based diffusion''} in the final manuscript. Both terms refer to the exact same architectural classification and evaluation setting.

\section{User study}
\label{appx:user_study}
To further evaluate the quality of the generated motion, we conducted a user study to observe the subjective quality of the generated motion using our method and baselines.
We asked $85$ participants to rank $11$ motion sequences from our method and baselines, based on quality of motion, alignment to the input sentence, naturalness, and smoothness.
The four methods of segment-based and volume diffusion are \textit{ours-28}, \textit{MD-28}, \textit{ours-196}, and motion diffusion.

We quantified the user study with two preference scores; the first to evaluate the extended generation (\textit{ours-28} and \textit{MD-28}) participant rate the overall quality of the motion in terms of naturalness (from 1 = ``Very unnatural" to 5 = ``Very natural"), motion smoothness (from 1 = ``Very smooth" to 5 = ``Very natural"), and alignment with the input sentence (from 1 = ``Not at all" to 5 = ``Perfectly aligned").
Then, we scaled each to $0$ and $1$ and inverted them.
The second asks the participants to compare the motions generated by our methods and baselines (\textit{ours-28}, \textit{MD-28}, \textit{ours-196} and MotionDiffuse) using the choices (``A", ``B", ``Both look similar to me").
We observe that our method has a preference score of $\sim 86\%$, $\sim 77\%$ and $\sim 80\%$ in naturalness, smoothness, and alignment to text input, respectively, as seen in \cref{fig:user_study}.

\section{Datasets}
The KIT-ML dataset~\cite{Plappert2016} provides $3911$ motion sequences and $6353$ sequence-level natural language descriptions. 
HumanML3D~\cite{Guo_2022_CVPR} re-annotates the AMASS dataset~\cite{AMASS_ICCV_2019} and the HumanAct12 dataset~\cite{guo2020action2motion}. 
It provides $44970$ annotations on $14616$ motion sequences.

\section{Evaluation Metrics}
We adhere to evaluation protocols from~\cite{zhang2022motiondiffuse, chen2023executing, zhang2025motion, light_t2m_2025}, assessing methods with four metrics (R-Precision, Fréchet Inception Distance (FID), Multi-modal Distance and Multimodality), each repeated $20$ times with a $95\%$ confidence interval.
\begin{enumerate}
    \item \textbf{R-Precision}: measures motion-to-text retrieval accuracy by randomly selecting 31 mismatched text prompts and reporting top-k accuracies based on embeddings. 
    \item \textbf{Fréchet Inception Distance (FID)}: evaluates realism and diversity of motions by comparing distributions of embeddings from generated and ground-truth data via a pretrained motion encoder. 
    \item \textbf{Multi-modal Distance}: uses a pretrained model to embed motions and calculates Euclidean distance between text and motion features. 
    \item \textbf{Multimodality}: assesses the diversity of motions from a single text by measuring joint position differences across $32$ generated sequences. 
\end{enumerate}

\section{Normalizing Flows}
\label{appx:rnf}
We use a real-valued non-volume preserving (Real-NVP) \cite{dinh2017density} as a probabilistic transformation, as depicted in the following equation:
\begin{equation}
    \begin{aligned}
        &[\sigma_{\phi}, \mu_{\phi}] = f_{\phi}(x^{i-1}_0, \varphi(u^{(j)})) \\
        &x^i_0 = x^{i-1}_0 \odot \sigma_{\phi}(x^{i-1}_0, \varphi(u^{(j)}))   +\mu_{\phi}(x^{i-1}_0, \varphi(u^{(j)}))
    \end{aligned}
\end{equation} 
where $\odot$ is element-wise multiplication, and both $\sigma_{\phi}$ and $\mu_{\phi}$ are deep feed-forward networks.
During training, we inflate the input to the flow with a Gaussian noise with a small standard deviation for more robustness, as in Horvat \etal~\cite{DNF_NEURIPS21}.
\begin{equation}
    \begin{aligned}
        \Tilde{x}^{i-1}_0 &= x^{i-1}_0 + \epsilon_i \text{ where } \epsilon_i  \sim \mathcal{N}(0, \sigma_{inf} I) \\
        [\sigma_{\phi}, \mu_{\phi}] &= f_{\phi}(\Tilde{x}^{i-1}_0, \varphi(u^{(j)})) \\
        x^i_0 &= x^{i-1}_0 \odot \sigma_{\phi}(\Tilde{x}^{i-1}_0, \varphi(u^{(j)})) \\ 
        &\quad + \mu_{\phi}(\Tilde{x}^{i-1}_0, \varphi(u^{(j)}))
    \end{aligned}
\end{equation}

\subsection{Numerical instability}
\label{appx:nf_nistability}
While we attempted to adapt normalizing flows for temporal modelling by augmenting them with an LSTM, they still suffer from numerical instability, particularly during multi-step inverse flow computations. 
The primary source of this instability is the $\sigma_{\phi}$ function within the mapping, where large values can cause an overflow of the inverse mapping $f^{(-i)}_{\phi}$, resulting in a NaN in the loss function.

We mitigated this by clamping the values to an acceptable range using a HardTanh function with limits of $- 0.1 \text{ and } 0.1$. 
While this successfully resolved the instability in the loss, it did not address the issue of very long recurrence. 
We identify this as a limitation of our current method, and a more comprehensive analysis of this long-recurrence behaviour will be addressed in future work.

\begin{figure}[H]
    \centering
    \includegraphics[width=\linewidth]{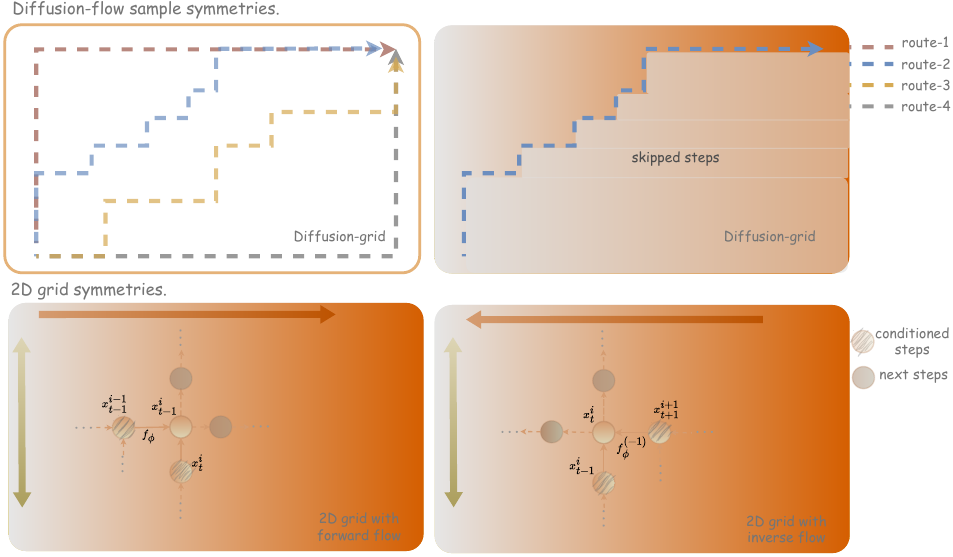}
    \caption{
    \textbf{Diffusion-flow symmetries.} By design, we can arrive at \textit{diffusion-flow} samples following different routes. 
    Some routes are more efficient than others as they allow for skipping diffusion sampling steps, but all are mathematically equivalent.
    \textbf{2D grid symmetries.} Normalizing flows properties gives us two grids (\textit{forward-flow} and \textit{inverse-flow}).
    Both are mathematically equivalent, but the \textit{forward-flow} is more numerically stable.
    }
    \label{fig:flow_properties}
\end{figure}

\section{Notes on recurrent formulation properties}
\label{appx:rfdm_properties}
The design of our 2D grid exhibits inherent symmetries, which can be exploited to skip diffusion sampling steps, leading to more efficient inference. 
While this primarily benefits inference, a different configuration allows for efficient training, albeit with a trade-off in flexibility during the inference phase. 
This section will detail these properties and their implications for computational efficiency.

\paragraph{\textbf{Diffusion-flow sample symmetries.}}
By design, we can arrive at \textit{diffusion-flow} samples through multiple mathematically equivalent sampling paths (see~\cref{fig:flow_properties}).
Empirically, however, these paths yield varying sample quality and computational efficiency. 
Specifically, paths that rely on the normalizing flow for temporal sampling produce lower quality, as the flow has less capacity than the denoiser and is not as well-suited for temporal modelling. 
In terms of efficiency, paths that require fewer diffusion steps are more computationally efficient, particularly with larger sequence lengths, as the denoising process is the most expensive operation. 
This design, therefore, allows for an explicit trade-off between computational efficiency and sample quality.

\paragraph{\textbf{2D grid symmetries.}}
Since normalizing flows are invertible transformations, they allow for two distinct grid formulations for conditioning the reverse process: one with a forward flow and one with an inverse flow. We can thus condition the reverse process on either $x^{i+1}_{t-1}$ or $x^{i-1}_{t-1}$.
While both options are mathematically equivalent, conditioning on $x^{i-1}_{t-1}$ is more numerically stable as it avoids computing an additional flow inverse transformation. 
We adopted this latter approach to parameterize our reverse flow and the associated loss function, as will be described in the next section.

\section{RDM sampling}
\label{appx:diff_sample}
Unlike previous methods \cite{zhang2022motiondiffuse, chen2023executing, han2023amd, light_t2m_2025}, our approach does not require sampling Gaussian noise for the full sequence. 
Instead, we leverage the flow model to skip diffusion steps by sampling in a ``staircase" through our 2D grid. 
This significantly enhances the computational efficiency and throughput when dealing with temporal data (see~\cref{alg:rfdm_sample_staircase}).

The point at which we initiate this staircase sampling presents a unique trade-off between computational complexity and sampling quality. 
One extreme, which we term ``disentangled sampling", involves generating the first segment with \textit{diffusion-only} and then rolling out the rest using only the normalizing flow. 
This offers the lowest computational complexity but results in the lowest sample quality, as it effectively disentangles the diffusion and temporal axis.
The other extreme is to initiate the staircase from the beginning with our \textit{diffusion-flow} approach.
This path yields the highest computational complexity but also produces better sample quality. 
A comparison of these strategies is detailed in~\cref{tab:sampling_metrics} and ~\ref{tab:sampling_compute_cost}.

\subsection{Beyond horizon inference complexity}
\begin{figure*}[htbp]
     \centering
     \begin{minipage}[b]{0.48\textwidth}
         \centering
         \includegraphics[width=\textwidth]{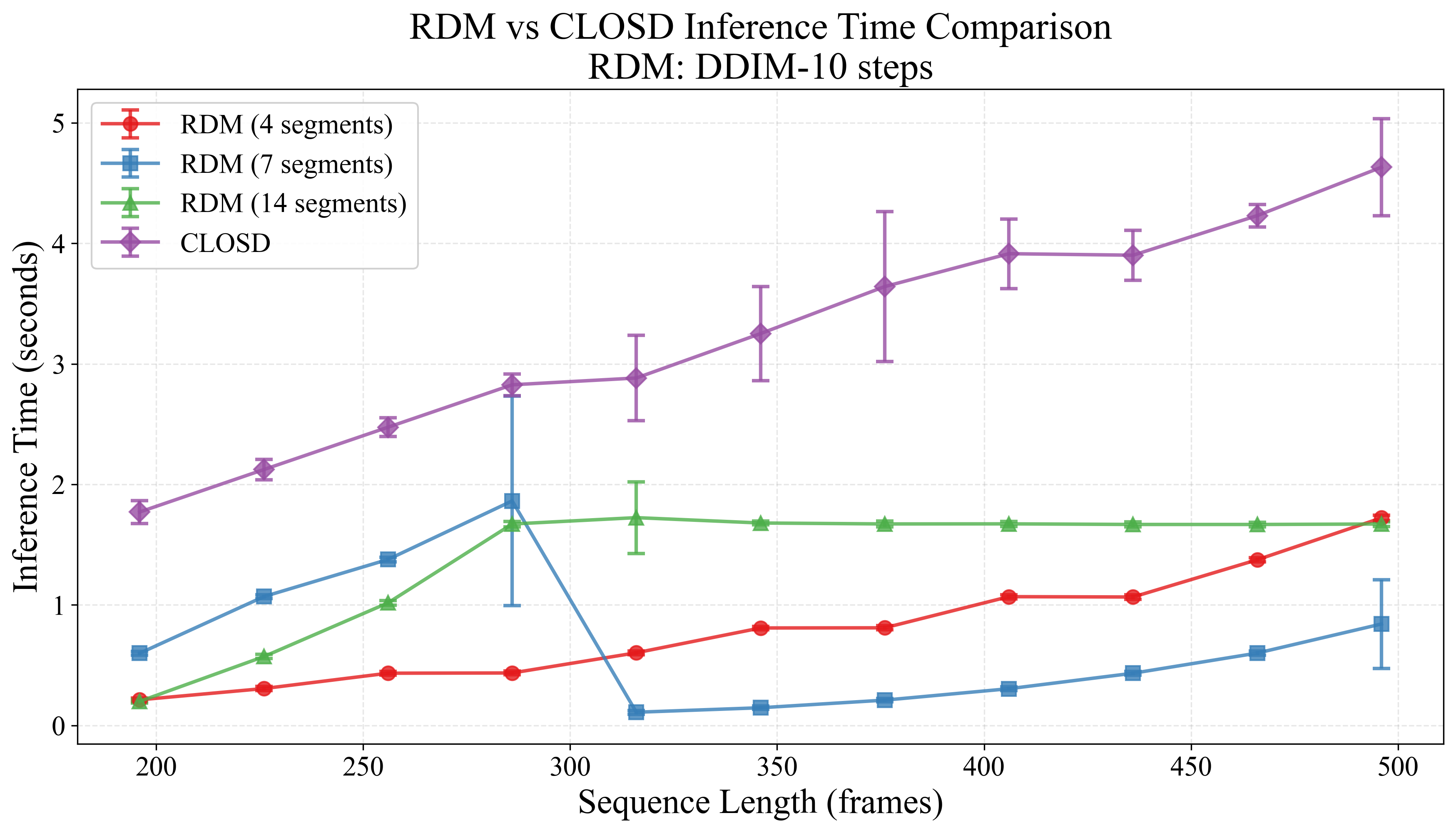}
         \label{fig:image_left_appendix}
     \end{minipage}
     \hfill
     \begin{minipage}[b]{0.48\textwidth}
         \centering
         \includegraphics[width=\textwidth]{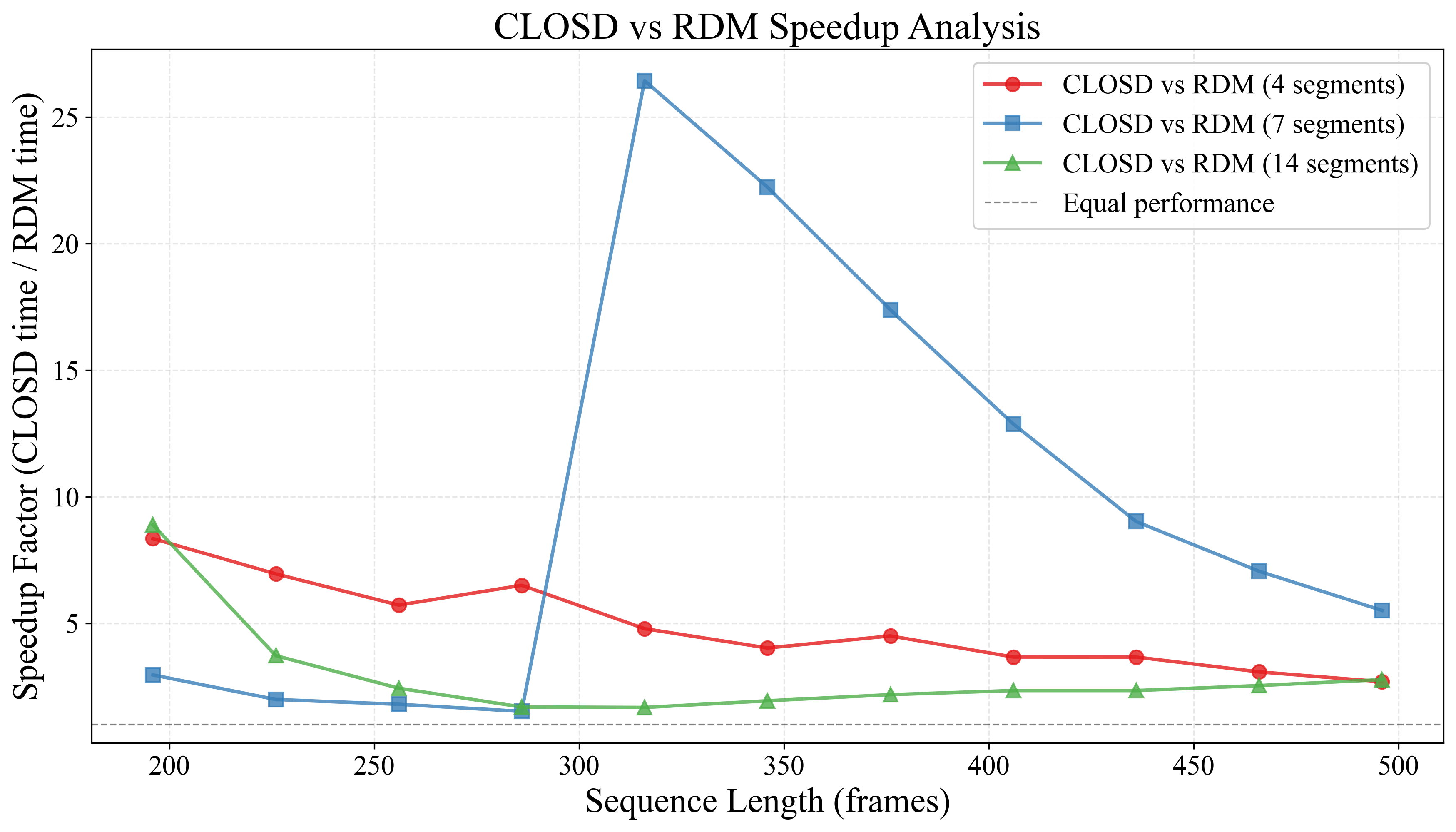}
         \label{fig:image_right_appendix}
     \end{minipage}
     
     \caption{
     \textbf{Computational efficiency comparison between RDM and CLoSD (DiP)~\cite{tevet2025closd}.} (left) Inference time versus sequence length for RDM with three segment numbers (4, 7, 14 segments, DDIM-10 steps) and CLoSD (DiP). Error bars represent $\pm3$ standard deviation.
     (right) Speedup analysis showing CLoSD (DiP) inference time divided by RDM inference time. 
     RDM consistently outperforms CLoSD (DiP) by a significant margin across all sequence lengths, with higher speedups for shorter sequences. 
     The speedup depends on the segment configuration: RDM-49 is more efficient than RDM-28, and RDM-14 is the least efficient.
     }
     \label{fig:rollout_compute_comparision_ddim10}
\end{figure*}

\begin{figure*}[htbp]
     \centering
     \begin{minipage}[b]{0.48\textwidth}
         \centering
         \includegraphics[width=\textwidth]{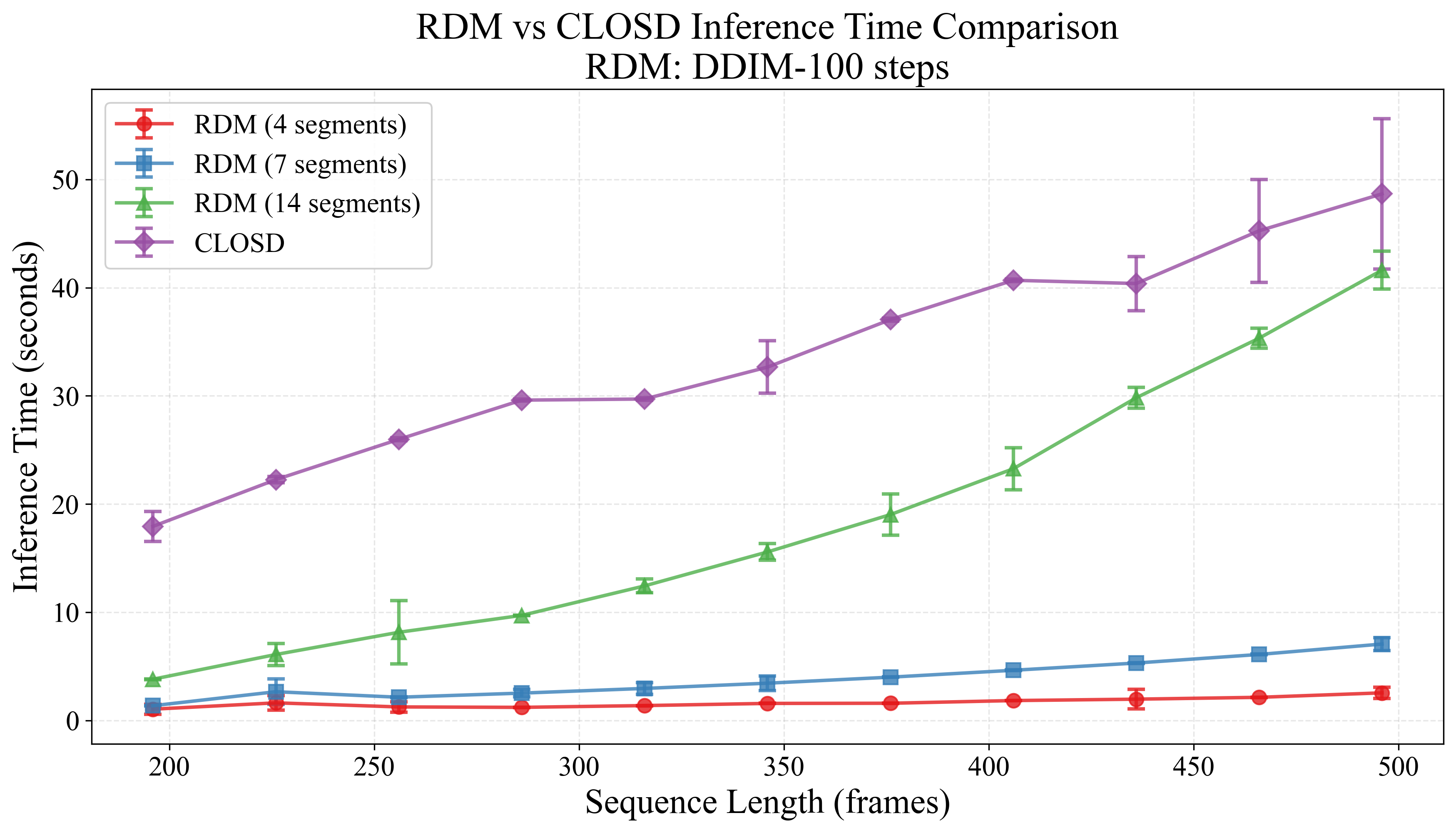}
         \label{fig:image_left_ddim100}
     \end{minipage}
     \hfill
     \begin{minipage}[b]{0.48\textwidth}
         \centering
         \includegraphics[width=\textwidth]{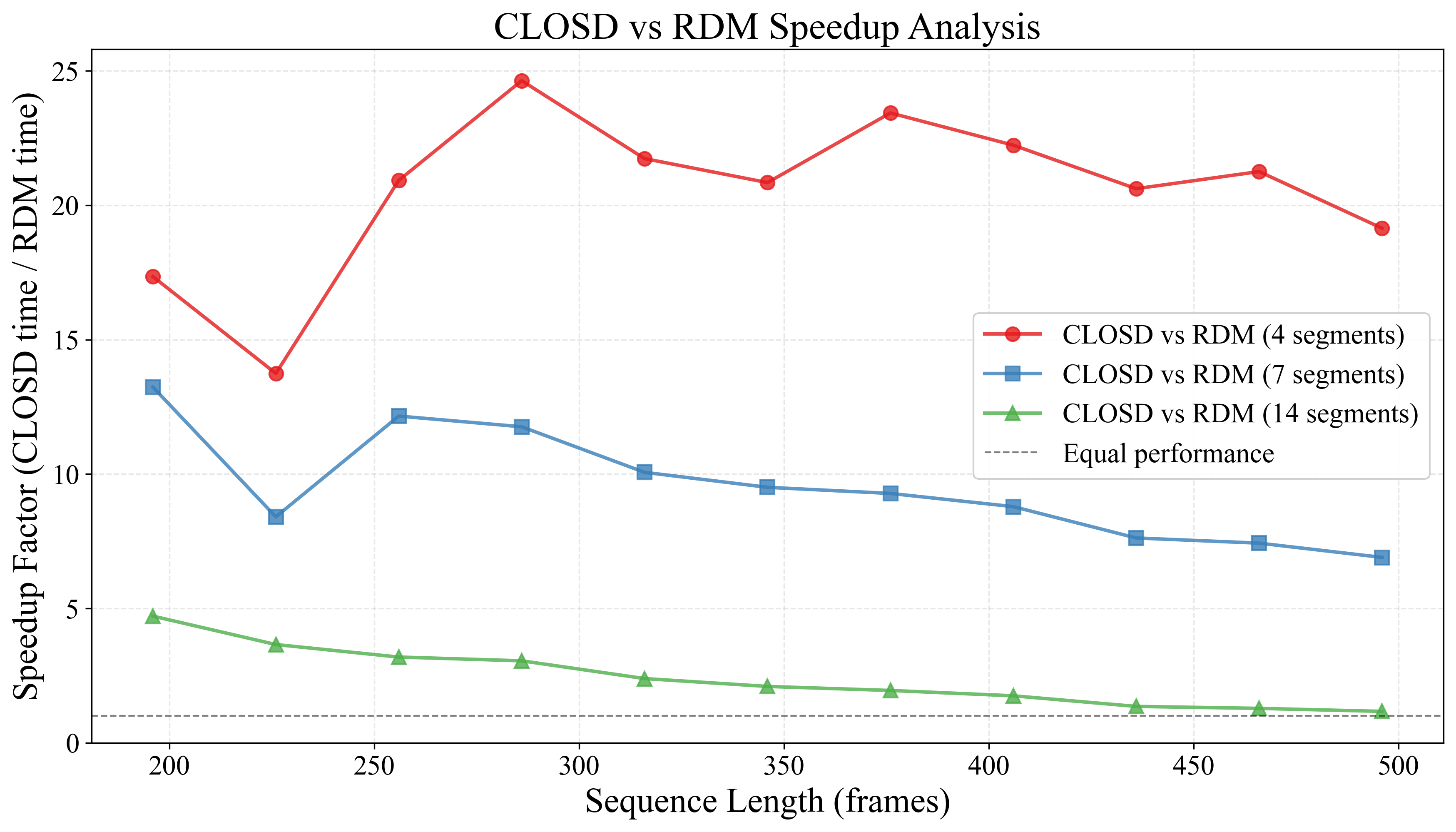}
         \label{fig:image_right_ddim100}
     \end{minipage}
     
     \caption{
     \textbf{Computational efficiency comparison between RDM and CLoSD (DiP)~\cite{tevet2025closd}.} (left) Inference time versus sequence length for RDM with three segment numbers (4, 7, 14 segments, DDIM-100 steps) and CLoSD (DiP). Error bars represent $\pm3$ standard deviation.
     (right) Speedup analysis showing CLoSD (DiP) inference time divided by RDM inference time. 
     RDM consistently outperforms CLoSD (DiP) by a significant margin across all sequence lengths, with higher speedups for shorter sequences. 
     The speedup depends on the segment configuration: RDM-49 is more efficient than RDM-28, and RDM-14 is the least efficient.
     }
     \label{fig:rollout_compute_comparision_ddim100}
\end{figure*}

To better understand the impact of sampling steps and model configuration on inference beyond the training horizon ($196$ frames), we compare three different \shortmethod configurations (4, 7, and 14 segments) with varying DDIM sampling steps against CLoSD~\cite{tevet2025closd}, in \cref{fig:rollout_compute_comparision_ddim100} and \ref{fig:rollout_compute_comparision_ddim10}. 
\shortmethod consistently proves more efficient than CLoSD~\cite{tevet2025closd} (shown on the left) and offers a significant speed advantage over CloSD~\cite{tevet2025closd} (shown on the right). 
However, this speed advantage depends on the segment configuration and sequence length, with \shortmethod-49 being the most efficient, followed by \shortmethod-28 and \shortmethod-14, which is the least efficient.

The computational complexity of all configurations \shortmethod-49/28/14 rises with an increasing number of segments, with \shortmethod-14 experiencing a nonlinear growth (DDIM-100, see \cref{fig:rollout_compute_comparision_ddim100}). 
However, in \cref{fig:rollout_compute_comparision_ddim10} (DDIM-10), we see an increase, but around $280$ it stabilises and maintains a steady speed. 
This is because we begin sampling only with the normalising flow (``disentangled sampling") for the remaining frames. 
Note that we start staircase sampling at the denoising step equal to the number of segments, and if it is not specified as in DDIM-10, we commence at the first step. 
Based on the rate of increase, if we extrapolate further and generate longer sequences without skipping steps, \shortmethod will eventually reach CLoSD time complexity.

\begin{table*}[t]
\resizebox{\textwidth}{!}{\begin{tabular}{llllllll}
\hline
\multicolumn{1}{c}{\multirow{2}{*}{Method}} & \multicolumn{3}{c}{R-Precision ($\uparrow$)} & \multirow{2}{*}{FID ($\downarrow$)} & \multirow{2}{*}{MultiModal Dist ($\downarrow$)} & \multirow{2}{*}{MultiModality ($\uparrow$)} \\ \cline{2-4}
\multicolumn{1}{c}{}                        & Top 1     & Top 2    & Top 3    &                      &                                  &                            &                                \\ \hline
Real motions                                & $0.5138$ & $0.6957$ & $0.7937 $ & $0.002 ^{\pm.000}$ & $2.974 ^{\pm.008}$ & - \\ \hline
Disentangle    & $0.3022^{\pm.0091}$ & $0.4954^{\pm.0084}$ & $0.6161^{\pm.0063}$ & $6.2045^{\pm.1114}$ & $4.4230^{\pm.05}$ & \bstcell$1.6361^{\pm.1407}$ \\
Entangle-996      & \bstcell$0.405^{\pm.002}$ & \bstcell$0.593^{\pm.003}$ & \bstcell$0.702^{\pm.002}$ & \bstcell$0.402^{\pm.018}$ & \midcell$3.651^{\pm.008}$ & $1.199^{\pm.04}$ \\
Entangle-900      & \worcell$0.334^{\pm.012}$ & \worcell$0.537^{\pm.005}$ & \worcell$0.667^{\pm.01}$ & \worcell$3.824^{\pm.233}$ & \bstcell$2.761^{\pm.019}$ & \midcell$1.284^{\pm.078}$ \\
Entangle-800      & \midcell$0.335^{\pm.005}$ & \midcell$0.539^{\pm.006}$ & \midcell$0.674^{\pm.002}$ & \midcell$3.648^{\pm.221}$ & \worcell$3.821^{\pm.06}$ & \worcell$1.241^{\pm.115}$ \\ \hline
\end{tabular}}
\caption{
\textbf{Quantitative results on the KIT-ML \cite{Plappert2016} test set.}
All methods use ground truth motion length. Evaluations were conducted $20$ times with $95\%$ CI reported. 
Rows are colour-coded as \bsttxt{best}, \midtxt{second best}, and \wortxt{third best}.
RDM’s staircase sampling trade-off between computational complexity and sampling quality. 
We found that starting the sampling at segment length provides the best result compared to starting earlier (Entangle-x, where x indicates the staircase sampling start) or ``disentangled", where we sample the first segment with \textit{diffusion-only} and roll out the rest with \textit{flow-only}. 
We use \shortmethod-28 for all experiments.
}
\label{tab:sampling_metrics}
\end{table*}

\begin{table*}[t]
\centering
\begin{tabular}{lcllcllc}
\hline
\multicolumn{1}{c}{\multirow{3}{*}{Method (\shortmethod-28)}} &
  \multicolumn{1}{l}{Total Inference Time (s) $\downarrow$} &
  \multicolumn{1}{c}{FLOPs (T) $\downarrow$} &
  \multirow{3}{*}{Parameters} \\ \cline{2-3}
\multicolumn{1}{c}{} &
  DDPM &
  DDPM &
   &
   \\ \hline
\multicolumn{1}{c}{} &
  1000 &
  1000 &
   &
   \\ \hline
Disentangle  & \bstcell$11.10^{\pm.59}$ & \bstcell$2.31$ & $x \in \mathbb{R}^{49 \times 251}$ \\
Entangle-997 & \midcell$11.7^{\pm2.02}$ & \midcell$2.32$ & $x \in \mathbb{R}^{49 \times 251}$ \\
Entangle-900 & \worcell$13.37^{\pm.31}$ & \worcell$4.06$ & $x \in \mathbb{R}^{49 \times 251}$ \\
Entangle-800 & $18.02^{\pm4.14}$ & $5.5$ & $x \in \mathbb{R}^{49 \times 251}$ \\ \hline
\end{tabular}
\caption{
\textbf{Inference time comparison.}
All methods use ground truth motion length. Evaluations were conducted $20$ times with $95\%$ CI reported. 
Rows are colour-coded as \bsttxt{best}, \midtxt{second best}, and \wortxt{third best}.
RDM’s staircase sampling trade-off between computational complexity and sampling quality. 
Although ``disentangled sampling" (sampling the first segment with \textit{diffusion-only} and the rest with \textit{flow-only}) has the lowest cost, it yields the worst results compared to Entangle-x, which starts staircase sampling at point x.
}
\label{tab:sampling_compute_cost}
\end{table*}

\section{Segment-based diffusion baselines}
\label{appx:md_basline}
\noindent{\textbf{Relation to latent-space diffusion.}} RDM and latent-space diffusion models improve efficiency along complementary axes: latent models reduce spatial dimensionality by compressing each frame, or the whole sequence, into a compact code, whereas RDM restructures long-range \emph{temporal} dependencies via its 2D recurrent grid, independent of the representation space used for individual frames. Because the two mechanisms are decoupled, RDM could in principle be combined with a latent-space denoiser for compounding efficiency gains; we leave this combination for future work.

To maintain flexibility and facilitate fair comparisons, we chose to denoise in pose space, similar to MotionDiffuse~\cite{zhang2022motiondiffuse}.
This decision was driven by the inherent limitations of latent diffusion methods, which require training a different VAE for each configuration of the number of segments, thereby complicating our ability to experiment with different recurrence lengths.

While our recurrent diffusion framework is agnostic to the choice of denoiser, we adopted the denoiser backbone from MotionDiffuse~\cite{zhang2022motiondiffuse}. 
We chose this for its excellent performance and its widespread use in other baselines~\cite{han2023amd, shi2024amdm, Zhang_2023_ICCV}, which establishes MotionDiffuse as our closest comparison point.

For a rigorous evaluation, we trained custom MotionDiffuse~\cite{zhang2022motiondiffuse} baselines to handle variable sequence horizons, mirroring our training setup. 
This baseline, which takes corrupted motion segments and predicts the next clean segments, was trained in two settings: one with $4$ segments (MD-49) and another with $7$ segments (MD-28). 
These MD-x baselines effectively represent RDM's architecture without recurrent connections, with diffusion applied to each segment independently. This effectively makes MD-x serve as an ablation of RDM's recurrent connection.

\section{Implementation details}
\label{appx:implement_details}
\paragraph{\textbf{Training settings.}}
We train RDM using MoMo-Adam optimizer ~\cite{Schaipp2023} with a learned rate $1e^{-4}$ for the flow and $2e^{-4}$ learning rate for the diffusion denoiser model, with a batch size $512$.
Our full training takes $\sim 4.5$ days on 2 H100 Nvidia GPUs.
We train the full model for $50$ epochs in total, and we artificially increased the data length by a factor of $25$ for each GPU used.
We train the flow and denoiser together to minimize the reconstruction loss (see~\cref{alg:diffusion_update}).

For inference-time comparison and computational efficiency, we ran all evaluations at a batch size of $ 128$ and a sequence length of $196$ $20$ times, reporting the mean inference time and standard deviation. 
For the FLOPs, we use the \textit{calflops} Python package to calculate the FLOPs on a batch size of $1$ with $196$ sequence length.
We ran both experiments on a single NVIDIA H100 GPU.

\paragraph{\textbf{Architectures.}}
We use $4/7$ segments for a maximum sequence length of $196$, yielding a segment length of $49/28$ frames.
For normalizing flow, we used a $1$ layer LSTM with $256$ hidden dimensions, and for the invertible transformation, we used $256$ feature dimensions and $6$ transformation blocks.
For the diffusion denoiser, we utilise a similar temporal transformer-based architecture to that of MotionDiffuse~\cite{zhang2022motiondiffuse}.
We used $8$ transformer layers with $512$ latent dimensions and $8$ heads.
As for the text conditioning, we used a pre-trained CLIP ViT-B/32 model~\cite{pmlr_v139_radford21a} to encode text and then added four more transformer encoder layers. 
The latent dimension of the text encoder and the motion decoder are $256$ and $512$, respectively. 
As for the diffusion model, the number of diffusion steps is $1000$, and the variances $\beta_t$ are linearly from $0.0001$ to $0.02$.

To incorporate temporal conditioning, we use positional encoding with different frequencies for the diffusion step number and the frame number, each mapped into latent space with a separate shallow MLP.
We follow the same conditioning method as in MotionDiffuse~\cite{zhang2022motiondiffuse} and add the mapped latents of frame and time to the projected text embedding.
Then, we apply cross-attention in the transformer to predict the denoiser, as in MotionDiffuse~\cite{zhang2022motiondiffuse}.

\newpage
\raggedbottom
\section{Loss derivation (proof)}
\label{appx:loss_derivation}
We use the forward normalizing flow according to the 2D grid defined in the previous section \ref{appx:rfdm_properties}. 
As discussed in the main text, our forward diffusion process across the grid is defined by the following equations:
\begin{align}
    q(x^i_{1:T}| x^i_0, x^{i-1}_T) &= \prod_{t=1}^T q(x^i_t| x^i_{t-1}, x^{i-1}_t) \\
    q(x^{1:L}_{1:T}| x^0_0, x^{0-1}_T) &= \prod_{t=1}^T q(x^{1:L}_t| x^{1:L}_{t-1}) \\
    q(x^{1:L}_{1:T}| x^0_0, x^{0-1}_T) &= \prod_{i=1}^L \prod_{t=1}^T q(x^i_t| x^i_{t-1}, x^{i-1}_t)
\end{align}
As we are using a 2D grid with forward normalizing flow, our reverse diffusion process is defined by the next equations:
\begin{align}
    p_{\theta}(x^i_{1:T}| x^i_0, x^{i-1}_T) = \prod_{t=1}^T p_{\theta}(x^i_{t-1}| x^i_t, x^{i-1}_{t-1}) \\
    p_{\theta}(x^{0:L}_{0:T}) = p(x^{0:L}_T) \prod_{t=1}^T p_{\theta}(x^{0:L}_{t-1}| x^{0:L}_t) \\
    p_{\theta}(x^{0:L}_{0:T}) = p(x^0_T) \prod_{i=1}^L \prod_{t=1}^T p_{\theta}(x^i_{t-1}| x^i_t, x^{i-1}_{t-1})
\end{align}
Let us begin by equating our data log probability $\log(p_{\theta}(x^{0:L}_0))$ as follows:
\begin{align*}
   - \log(p_{\theta}(x^{0:L}_0)) &\leq - \log(p_{\theta}(x^{0:L}_0)) + D_{kl}(q(x^{0:L}_{1:T}|x^{0:L}_0) \| p_{\theta}(x^{0:L}_{1:T} | x^{0:L}_0)) \\
   &= - \log(p_{\theta}(x^{0:L}_0)) + \mathbb{E}_{x^{0:L}_{1:T} \sim q(x^{0:L}_{1:T} | x^{0:L}_0)} \big[ \log \frac{q(x^{0:L}_{1:T} | x^{0:L}_0)}{p_{\theta}(x^{0:L}_{0:T})/p_{\theta}(x^{0:L}_0)} \big] \\
   &= \mathbb{E}_{q} \big[ \log \frac{q(x^{0:L}_{1:T} | x^{0:L}_0)}{p_{\theta}(x^{0:L}_{0:T})} \big]
\end{align*}

Then let $\mathcal{L}_{VLB} = \mathbb{E}_{q(x^{0:L}_{0:T})} \big[ \log \frac{q(x^{0:L}_{1:T} | x^{0:L}_0)}{p_{\theta}(x^{0:L}_{0:T})}\big] \geq - \mathbb{E}_{q(x^{0:L}_{0:T})} p_{\theta}(x^{0:L}_0)$

\begin{align*}
    \mathcal{L}_{VLB} &= \mathbb{E}_{q} \big[ \log \frac{q(x^{0:L}_{1:T} | x^{0:L}_0)}{p_{\theta}(x^{0:L}_{0:T})}\big] \\
    &= \mathbb{E}_{q} \big[ \log \frac{\prod_{t=1}^T q(x^{0:L}_t | x^{0:L}_{t-1})}{p_{\theta}(x^{0:L}_T) \prod_{t=1}^T p_{\theta}(x^{0:L}_{t-1} | x^{0:L}_t)}\big] \\
    &= \mathbb{E}_{q} \big[ -\log(p_{\theta}(x^{0:L}_T)) + \sum_{t=1}^T \log \frac{ q(x^{0:L}_t | x^{0:L}_{t-1})}{p_{\theta}(x^{0:L}_{t-1} | x^{0:L}_t)}\big] \\
    &= \mathbb{E}_{q} \big[ -\log(p_{\theta}(x^{0:L}_T)) + \sum_{t=2}^T \log \frac{ q(x^{0:L}_t | x^{0:L}_{t-1})}{p_{\theta}(x^{0:L}_{t-1} | x^{0:L}_t)} + \log \frac{ q(x^{0:L}_1 | x^{0:L}_0)}{p_{\theta}(x^{0:L}_0 | x^{0:L}_1)}\big]
\end{align*}

\begin{align*}
    \mathcal{L}_{VLB} &= \mathbb{E}_{q} \big[ -\log(p_{\theta}(x^{0:L}_T)) + \sum_{t=2}^T \log \big( \frac{ q(x^{0:L}_{t-1} | x^{0:L}_t, x^{0:L}_0)}{p_{\theta}(x^{0:L}_{t-1} | x^{0:L}_t)} . \frac{ q(x^{0:L}_t | x^{0:L}_0)}{q(x^{0:L}_{t-1} | x^{0:L}_0)} \big) \\ 
    \; \; & + \log \frac{ q(x^{0:L}_1 | x^{0:L}_0)}{p_{\theta}(x^{0:L}_0 | x^{0:L}_1)}\big] \; \; \text{(Bayes's rule)} \\
    &= \mathbb{E}_{q} \big[ -\log(p_{\theta}(x^{0:L}_T)) + \sum_{t=2}^T \log \frac{ q(x^{0:L}_{t-1} | x^{0:L}_t, x^{0:L}_0)}{p_{\theta}(x^{0:L}_{t-1} | x^{0:L}_t)} + \sum_{t=2}^T \log \frac{ q(x^{0:L}_t | x^{0:L}_0)}{q(x^{0:L}_{t-1} | x^{0:L}_0)} + \log \frac{ q(x^{0:L}_1 | x^{0:L}_0)}{p_{\theta}(x^{0:L}_0 | x^{0:L}_1)}\big] \\
    &= \mathbb{E}_{q} \big[ -\log(p_{\theta}(x^{0:L}_T)) + \sum_{t=2}^T \log \frac{ q(x^{0:L}_{t-1} | x^{0:L}_t, x^{0:L}_0)}{p_{\theta}(x^{0:L}_{t-1} | x^{0:L}_t)} + \log \frac{ q(x^{0:L}_T | x^{0:L}_0)}{q(x^{0:L}_1 | x^{0:L}_0)} + \log \frac{ q(x^{0:L}_1 | x^{0:L}_0)}{p_{\theta}(x^{0:L}_0 | x^{0:L}_1)}\big] \\ &\hspace{1cm} \text{(Intermediate terms cancels)} \\
    &= \mathbb{E}_{q} \big[ \log \frac{q(x^{0:L}_T | x^{0:L}_0)}{p_{\theta}(x^{0:L}_T)} + \sum_{t=2}^T \log \frac{ q(x^{0:L}_{t-1} | x^{0:L}_t, x^{0:L}_0)}{p_{\theta}(x^{0:L}_{t-1} | x^{0:L}_t)} - \log (p_{\theta}(x^{0:L}_0 | x^{0:L}_1))\big] \; \; \text{(Rearranging logs)} \\
    &= \mathbb{E}_{q} \big[ \underbrace{D_{kl}(q(x^{0:L}_T | x^{0:L}_0) \| p_{\theta}(x^{0:L}_T))}_{\mathcal{L}^{0:1}_T} + \sum_{t=2}^T \underbrace{D_{kl}(q(x^{0:L}_{t-1} | x^{0:L}_t, x^{0:L}_0) \| p_{\theta}(x^{0:L}_{t-1} | x^{0:L}_t))}_{\mathcal{L}^{0:L}_{t-1}} \\
    & + \mathbb{E}_{q} [ \underbrace{ - \log (p_{\theta}(x^{0:L}_0 | x^{0:L}_1))}_{\mathcal{L}^{0:L}_0} ] \big] \\
\end{align*}

Let us label each term in the lower bound separately and further expand them:
\begin{align*}
    \mathcal{L}^{0:1}_T &= D_{kl}(q(x^{0:L}_T | x^{0:L}_0) \| p_{\theta}(x^{0:L}_T)) = \mathbb{E}_{q}[ \log \big( \frac{q(x^0_T | x^0_0) \prod_{i=1}^L q(x^i_T | x^i_0, x^{i-1}_T)}{p_{\theta}(x^0_T) \prod_{i=1}^L p_{\theta}(x^i_T|x^{i-1}_T)} \big) ] \\
    &= \mathbb{E}_{q} \big[ \log \frac{q(x^0_T | x^0_0)}{p_{\theta}(x^0_T)} + \sum_{i=1}^L \log \frac{q(x^i_T | x^i_0, x^{i-1}_T)}{p_{\theta}(x^i_T|x^{i-1}_T)} \big] \\
    &= \underbrace{D_{kl}(q(x^0_T | x^0_0) \| p_{\theta}(x^0_T))}_{L^0_T} + \sum_{i=1}^L \underbrace{D_{kl}(q(x^i_T | x^i_0, x^{i-1}_T) \| p_{\theta}(x^i_T|x^{i-1}_T))}_{L^i_T} \\
\end{align*}

\begin{align*}
    \mathcal{L}^{0:L}_0 &= - \log (p_{\theta}(x^{0:L}_0 | x^{0:L}_1)) = - \log (p_{\theta}(x^0_0 | x^0_1)) - \log (\prod_{i=1}^L p_{\theta}(x^i_0 | x^{i-1}_0, x^i_1)) \\
    &= \underbrace{ - \log (p_{\theta}(x^0_0 | x^0_1))}_{\mathcal{L}^0_0} + \sum_{i=1}^L \underbrace{ - \log ( p_{\theta}(x^i_0 | x^{i-1}_0, x^i_1))}_{\mathcal{L}^i_0}
\end{align*}

\begin{align*}
    \mathcal{L}^{0:L}_{t-1} &= D_{kl}(q(x^{0:L}_{t-1} | x^{0:L}_t, x^{0:L}_0) \| p_{\theta}(x^{0:L}_{t-1} | x^{0:L}_t)) = \log \frac{ q(x^{0:L}_{t-1} | x^{0:L}_t, x^{0:L}_0)}{p_{\theta}(x^{0:L}_{t-1} | x^{0:L}_t)} \\
    &= \log \big(\frac{ q(x^0_{t-1} | x^0_t, x^0_0) \prod_{i=1}^L q(x^i_{t-1} | x^i_t, x^{i-1}_{t-1}, x^i_0)}{p_{\theta}(x^0_{t-1} | x^0_t) \prod_{i=1}^L p_{\theta}(x^i_{t-1} | x^i_t, x^{i-1}_{t-1})} \big) \\ 
    &= \mathbb{E}_{q} [ \log \frac{q(x^0_{t-1} | x^0_t, x^0_0)}{p_{\theta}(x^0_{t-1} | x^0_t)} + \sum_{i=1}^L \log \frac{q(x^i_{t-1} | x^i_t, x^{i-1}_{t-1}, x^i_0)}{p_{\theta}(x^i_{t-1} | x^i_t, x^{i-1}_{t-1})} ] \\
    &= \underbrace{D_{kl}(q(x^0_{t-1} | x^0_t, x^0_0) \| p_{\theta}(x^0_{t-1} | x^0_t))}_{\mathcal{L}^0_{t-1}} + \sum_{i=1}^L \underbrace{D_{kl}(q(x^i_{t-1} | x^i_t, x^{i-1}_{t-1}, x^i_0) \| p_{\theta}(x^i_{t-1} | x^i_t, x^{i-1}_{t-1}))}_{\mathcal{L}^i_{t-1}}
\end{align*}

Our final variational lower bound $\mathcal{L}_{VLB}$ takes the form:

\begin{equation}
    \begin{aligned}
    \mathcal{L}_{VLB} &= \sum_{i=0}^L \sum_{t=0}^T \mathcal{L}^i_t = \sum_{i=0}^L [ \mathcal{L}^i_T + \sum_{i=2}^T \mathcal{L}^i_{t-1} + \mathcal{L}^i_0 ] \\
    &= D_{kl}(q(x^0_T | x^0_0) \| p_{\theta}(x^0_T)) + \sum_{t=2}^T D_{kl}(q(x^0_{t-1} | x^0_t, x^0_0) \| p_{\theta}(x^0_{t-1} | x^0_t)) - \log (p_{\theta}(x^0_0 | x^0_1) \\
    & + \sum_{i=1}^L \big[ D_{kl}(q(x^i_T | x^i_0, x^{i-1}_T) \| p_{\theta}(x^i_T|x^{i-1}_T)) + \sum_{t=2}^T D_{kl}(q(x^i_{t-1} | x^i_t, x^{i-1}_{t-1}, x^i_0) \| p_{\theta}(x^i_{t-1} | x^i_t, x^{i-1}_{t-1})) \\
    & - \log ( p_{\theta}(x^i_0 | x^{i-1}_0, x^i_1))) \big]
    \end{aligned}
\end{equation}

The resulting training procedure is summarized in Algorithm~\ref{alg:diffusion_update} in the main text (\cref{sec:method}).

\subsection{Parameterization of training Loss.}
Each $\mathcal{L}^{.}_{T}$ is constant with respect to the diffusion encoder and can be ignored because $q$ has no learnable parameters and $x^0_T$ is Gaussian noise, while $x^{1:}_T$ are defined with the flow, which is only updated with the $x^{.}_0$.
Every KL term in $\mathcal{L}^{0}_{.}$ except $\mathcal{L}^{.}_{0}$ compares two Gaussians and is computed in closed form.
The remaining are the KL terms in $\mathcal{L}^{1:}_{.}$ except $\mathcal{L}^{1:}_{0}$. 
Each compares two unknown distributions and is defined using the normalizing flow.

\begin{align*}
    \mathcal{L}^i_{t-1} &= D_{KL} (q_{\phi}(x^i_{t-1} |x^i_t, x^{i-1}_{t-1}, x^i_0) \| p_{\theta}(x^i_{t-1} |x^i_t, x^{j-1}_{t-1})) \\
    \mathcal{L}^i_{t-1} &= \sum_{x^i_{t-1}} q^i_{\phi}(x^i_{t-1} |x^i_t, x^{i-1}_{t-1}, x^i_0) [\log \frac{q_{\phi}(x^i_{t-1} |x^i_0, x^0_0)}{p_{\theta}(x^i_{t-1} |x^i_t, x^{i-1}_{t-1})}]
\end{align*}

\subsubsection{The case of linear flow}
\label{appx:linear_flow}
As we restrict our flow to a linear transformation, the resulting distributions remain Gaussian.
All KL terms in $\mathcal{L}^{1:}_{.}$ are computed in closed form, with mean shifted and standard deviation scaled according to the normalizing flow.
Depending on the parametrization, we arrive at the same loss as Ho \etal~\cite{NEURIPS2020_4c5bcfec} with different weight terms.

\begin{align}
     \mathcal{L}^i_{t-1} &= D_{KL} (\mathcal{N}(x^i_{t-1}; \mu(x^i_t, x^{i-1}_{t-1}, x^i_0),\Sigma_t) \| \mathcal{N}(x^i_{t-1};\mu_{\theta}(x^i_{t},x^{i-1}_{t-1},t,i), \Sigma_{\theta})) \\
    \mathcal{L}_{t-1}^{i} &= \mathbb{E}_{q} \left[\Tilde{w}(t) \| x^i_0 - f_{\theta}(x^i_t, x^{i-1}_{t-1},t,i)\|^2\right] \; \; \text{(Same as diffusion loss)}
\end{align}

Where $\Tilde{w}(t)$ represents the signal-to-noise ratio, accounting for the flow transformation as well, and can take different forms depending on the chosen parameterization.
Here we show the case of the parametrization in Ho \etal~\cite{NEURIPS2020_4c5bcfec}, but other parametrizations cases are easily obtained:

\begin{align*}
    \Tilde{w}(t) = \frac{\beta^2_t}{2\sigma^2_t\alpha_t(1-\Bar{\alpha})} |det \frac{\partial f_{\phi}}{\partial x^i_0}|^{-1}
\end{align*}

\subsubsection{The generic case of non-linear flow}
Since we employ a generic nonlinear flow, each KL term in $\mathcal{L}^{1:}_{.}$ (except $\mathcal{L}^{1:}_{0}$) compares two unknown distributions defined via a normalizing flow.

Learning the posterior of these distributions typically requires techniques like Markov Chain Monte Carlo (MCMC)~\cite{Metropolis1953, Vono20}. However, MCMC methods suffer from slow convergence, making learning computationally infeasible across $1000$ diffusion steps.
This raises a crucial question: how can we efficiently learn the reverse process?

To bypass MCMC bottlenecks, we leverage the geometric properties of normalizing flows to derive an exact, closed-form solution for these KL terms through a three-step mapping procedure:
\begin{enumerate}
    \item \textbf{Inverse Flow Mapping:} We apply the inverse flow to map the complex samples $x^i_t$ back to the baseline, \textit{diffusion-only} state $x^0_t$.
    \item \textbf{Gaussian Evaluation:} In this \textit{diffusion-only} latent space, the distributions collapse to standard Gaussians, allowing us to compute the KL divergence analytically.
    \item \textbf{Temporal Forward Push:} We push the calculated distribution back to temporal step $i$ via the forward flow.
\end{enumerate}
Ultimately, this mapping transforms the complex objective into a simple KL divergence between two normal distributions, scaled by additional weighting terms derived from the inverse Jacobian determinant of the flow transformation:

\begin{align*}
     \log \big(q_{\phi}(x^i_{t-1} |x^i_t, x^{i-1}_{t-1}, x^i_0)\big) &= \log \mathcal{N}(x^0_{t-1}) - \sum_{j=1}^i \log |det \frac{\partial f_{\phi}}{\partial x^0_0}| \; \; \text{(Flow properties)} \\
     \log \big(p_{\theta}(x^i_{t-1} |x^i_t, x^{j-1}_{t-1})\big) &= \log \mathcal{N}(f^{-i}(x^i_{t-1})) - \sum_{j=1}^i \log |det \frac{\partial f_{\phi}}{\partial x^i_0}| \; \; \text{(Flow properties)} \\
     \mathcal{L}^i_{t-1} &= \sum_{x} q_{\phi}(x^0_{t-1} |x^0_t, x^0_0) \log \frac{\mathcal{N}(x^0_{t-1}; \mu(x^0_{t-2},x^0_0),\beta_t)}{\mathcal{N}(f^{-i}(x^i_{t-1});\mu_{\theta}(x^i_{t},x^{i-1}_{t-1},t,i), \Sigma_{\theta})} \big|det \frac{\partial f_{\phi}}{\partial x^i_0}\big|^{-1} \\ 
    &\hspace{1cm} \text{(Flow properties)}
\end{align*}

The KL between two Gaussians is computed in closed form, and following a similar parametrization to Ho \etal~\cite{NEURIPS2020_4c5bcfec}, the loss is further reduced to the difference between means, as depicted in the following equation:

\begin{align}
   \mathcal{L}_t^{i} &= \mathbb{E}_{q^0} [w(t) |det \frac{\partial f_{\phi}}{\partial x^i_0}|^{-1}\| \mu(x^i_t, x^{i-1}_{t-1}, x^i_0) -\mu_{\theta}(x^i_{t},x^{i-1}_{t-1},t,i)\|^2] \; \; \text{(Same as diffusion loss)}
\end{align}

The question remains: how to get the means of the distribution at the \textit{diffusion-only}.
As the forward process follows a Markovian chain, the mean is simply the previously sampled noisy segment $x^0_{t-1}$.
We choose to predict the clean segment, and we get the predicted mean following the grid (using the inverse flow and the forward diffusion) as follows:

\begin{align}
    \mu(x^i_t, x^{i-1}_{t-1}, x^i_0) &= \tilde{\mu}_t\left(f^{(-i)}_{\phi}(x^i_t), x^0_0\right) \\
    \mu_{\theta}(x^i_t, x^{i-1}_{t-1}, t, i) &= \tilde{\mu}_t\left(f^{(-i)}_{\phi}(x^i_t), \hat{x}^0_0\right) \\
    \mathcal{L}_{t-1}^{i} &= w_t \| x^0_0 -  \hat{x}^0_0 \|^2_2 \\
    \mathcal{L}_{t-1}^{i} &= w_t \| f^{(-i)}_{\phi}(x^i_0) -  f^{(-i)}_{\phi}(\hat{x}^i_0) \|^2_2 \label{eq:inv_mean_loss}
\end{align}

where $\tilde{\mu}_t(\cdot, \cdot)$ represents the analytical posterior mean of the diffusion process in the Gaussian baseline space, defined as:
\begin{equation}
    \tilde{\mu}_t(z_t, z_0) = \frac{\sqrt{\alpha_t}(1 - \bar{\alpha}_{t-1})}{1 - \bar{\alpha}_t} z_t + \frac{\sqrt{\bar{\alpha}_{t-1}}\beta_t}{1 - \bar{\alpha}_t} z_0
\end{equation}
By defining the inputs in terms of our inverted flow mappings—where $z_t = f^{(-i)}_{\phi}(x^i_t)$ and $z_0 = x^0_0$—we smoothly isolate the flow transformations. Cancelling out identical terms during the subtraction of $\mu$ and $\mu_{\theta}$ isolates the data reconstruction error, yielding the simplified loss formulations shown above.

We found that the previous loss~\cref{eq:inv_mean_loss} is unstable and difficult to optimize because it relies on the numerical stability of the inverse flow and assumes a perfect flow model. 
This makes the diffusion model's performance depend on the flow's effectiveness.
Therefore, we decided to optimize the subsequent loss instead.
As the flow is a smooth mapping, we can employ the Taylor expansion series around $0$, which leads us to the following equation:

\begin{align}
   \mathcal{L}_{t-1}^{i} &= \mathbb{E}_{q^0} [w(t) |det \frac{\partial f_{\phi}}{\partial x^i_0}|^{-1}\| x^i_0 - f_{\theta}(x^i_t, f^{i-1}_{t-1},t,i)\|^2] \; \; \text{(Same as diffusion loss)}
\end{align}

\end{document}